\documentclass[10pt,twocolumn,letterpaper]{article}

\usepackage{wacv}
\usepackage{times}
\usepackage{epsfig}
\usepackage{graphicx}
\usepackage{amsmath}
\usepackage{amssymb}
\usepackage{booktabs}

\usepackage{booktabs}
\usepackage{colortbl}
\usepackage{listings}
\usepackage{multirow}
\usepackage{pifont}
\usepackage{epsfig}
\usepackage{arydshln}
\usepackage{xcolor}
\usepackage[font=small,labelfont=bf,tableposition=top]{caption}
\usepackage{cuted}
\usepackage{float}

\newcommand{\cnn}{\mathrm{CNN}}
\newcommand{\wsi}{\mathrm{WSI}}
\newcommand{\wsis}{\mathrm{WSIs}}
\newcommand{\ssl}{\mathrm{SSL}}
\newcommand{\vit}{\mathrm{ViT}}

\newcommand{\scoremix}{\mathrm{ScoreMix}}
\newcommand{\score}{\mathrm{\textbf{ScoreMix}}}
\newcommand{\scorenet}{\mathrm{ScoreNet}}
\newcommand{\snet}{\mathrm{\textbf{ScoreNet}}}
\newcommand{\smix}{\mathrm{\textbf{ScoreMix}}}
\newcommand{\trois}{\mathrm{TRoIs}}
\newcommand{\roi}{\mathrm{RoI}}
\newcommand{\troi}{\mathrm{TRoIs}}
\newcommand{\TR}{\mathrm{\textbf{TRoIs}}}
\newcommand{\mil}{\mathrm{MIL}}
\newcommand{\he}{\mathrm{H \& E}}

\newcommand{\ck}{\textcolor{green!80!black}{\ding{51}}}
\newcommand{\xk}{\textcolor{red}{\ding{55}}}

\DeclareMathOperator*{\argmax}{arg\,max}

\newcommand*\rot{\rotatebox{90}}

\def\wacvPaperID{563} 

\wacvalgorithmstrack   

\wacvfinalcopy 

\ifwacvfinal
\usepackage[breaklinks=true,bookmarks=false]{hyperref}
\else
\usepackage[pagebackref=true,breaklinks=true,colorlinks,bookmarks=false]{hyperref}
\fi

\begin{document}

\title{ScoreNet: Learning Non-Uniform Attention and Augmentation for Transformer-Based Histopathological Image Classification}

\author{Thomas Stegm\"uller$^{1}$
\and
Behzad Bozorgtabar$^{1,2,3}$
\and 
Antoine Spahr$^{1}$
\and 
Jean-Philippe Thiran$^{1,2,3}$
\and $^{1}$EPFL, Switzerland \and $^{2}$CHUV, Switzerland \and $^{3}$CIBM, Switzerland 
\and {\tt\small \{firstname.lastname\}@epfl.ch}
}
\maketitle

\begin{strip}
    \centering
    \includegraphics[width=0.7\textwidth]{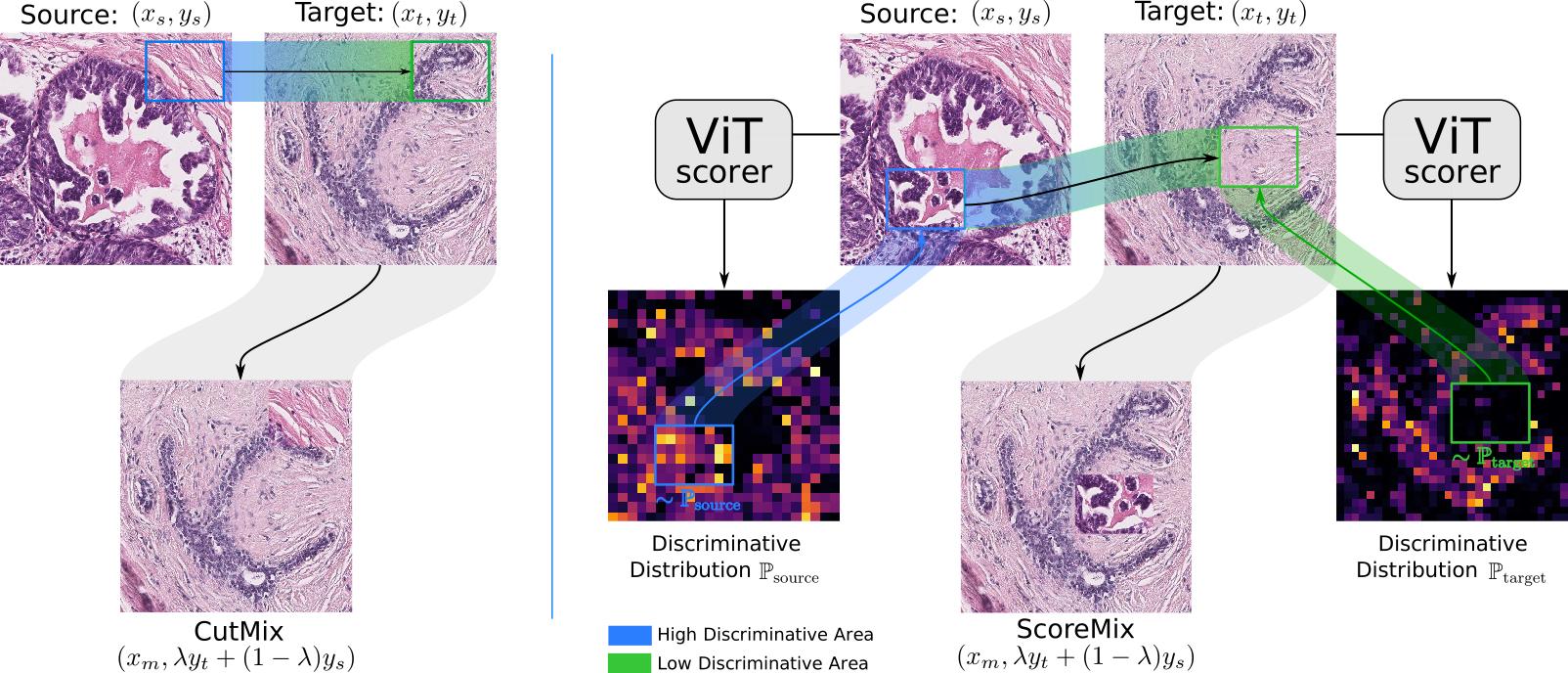}
    \captionof{figure}{CutMix (\textbf{left}) randomly mixes samples, yielding label misallocation, while our $\scoremix$ (\textbf{right}) creates a coherent artificial training pair $\left(x_{m}, y_{m} \right)$ by pasting a region of high semantic content from the source image, $x_{s}$ at a non-discriminative region of the target image $x_{t}$, and obtains a convex combination of the labels.}
    \label{fig:scoreMixV2}
\end{strip}

\thispagestyle{empty}

\begin{abstract}
Progress in digital pathology is hindered by high-resolution images and the prohibitive cost of exhaustive localized annotations. The commonly used paradigm to categorize pathology images is patch-based processing, which often incorporates multiple instance learning $(\mil)$ to aggregate local patch-level representations yielding image-level prediction. Nonetheless, diagnostically relevant regions may only take a small fraction of the whole tissue, and current $\mil$-based approaches often process images uniformly, discarding the inter-patches interactions. To alleviate these issues, we propose $\scorenet$, a new efficient transformer that exploits a differentiable recommendation stage to extract discriminative image regions and dedicate computational resources accordingly. The proposed transformer leverages the local and global attention of a few dynamically recommended high-resolution regions at an efficient computational cost. We further introduce a novel mixing data-augmentation, namely $\scoremix$, by leveraging the image’s semantic distribution to guide the data mixing and produce coherent sample-label pairs. $\scoremix$ is embarrassingly simple and mitigates the pitfalls of previous augmentations, which assume a uniform semantic distribution and risk mislabeling the samples. Thorough experiments and ablation studies on three breast cancer histology datasets of Haematoxylin \& Eosin $(\he)$ have validated the superiority of our approach over prior arts, including transformer-based models on tumour regions-of-interest $(\trois)$ classification. $\scorenet$ equipped with proposed $\scoremix$ augmentation demonstrates better generalization capabilities and achieves new state-of-the-art (SOTA) results with only 50\% of the data compared to other mixing augmentation variants. Finally, $\scorenet$ yields high efficacy and outperforms SOTA efficient transformers, namely TransPath~\cite{wang2021transpath} and SwinTransformer~\cite{liu2021swin}, with throughput around $3\times$ and $4\times$ higher than the aforementioned architectures, respectively.
\end{abstract}

\section{Introduction}
Due to the increasing availability of digital slide scanners enabling pathologists to capture high-resolution whole slide images $(\wsi)$, computational pathology is becoming a ripe ground for deep learning and recently witnessed a lot of advances. Nonetheless, the diagnosis from $\he$ stained $\wsis$ remains challenging. The difficulty of the task is a consequence of two inherent properties of histopathology image datasets: \textit{i)} the huge size for images and \textit{ii)} the cost of exhaustive localized annotations, making the usage of most deep learning models computationally infeasible. Patch-based processing approaches \cite{Srinidhi2021deep,mercan2019patch,hou2016patch} have become a \textit{de facto} practice for high dimensional pathology images that aggregate individual patch representation/classification predictions by, e.g., a convolutional neural network $(\cnn)$ for image-level prediction. Nonetheless, patch-based methods increase the requirement of patch-level labeling and further regions of interest $(\roi)$ detection as diagnostic-related tissue sections might only take a small fraction of the whole tissue, leading to considerable uninformative patches. Prior $\cnn$ methods \cite{ilse2018attention,li2021dual} have adopted multiple instance learning $(\mil)$ \cite{maron1998framework} to address the above issues, which incorporates an attention-based aggregation operator to identify tissue sub-regions of high diagnostic value automatically. Nonetheless, these MIL methods embed all the patches independently and discard the inter-patches correlation or only incorporate it at a later stage.

Recently, self-supervised learning $(\ssl)$ methods \cite{li2021dual,koohbanani2021self,srinidhi2021self,ciga2022self} aimed to construct semantically meaningful visual representations via pretext tasks for histopathological images. Despite their notable success using $\cnn$ backbones in improving classification performances, $\cnn$'s receptive field often restricts the learning of global context features. In another line of research, to compensate for the lack of diverse and large datasets, mixing augmentation techniques \cite{walawalkar2020attentive,yun2019cutmix,zhang2018mixup} have been developed to further enhance the performance of these models. While there have been substantial performance gains on natural image datasets, we argue that such data augmentations may not be helpful for histopathological images, as they risk creating locally ambiguous images or mislabelled samples.
Furthermore, contrary to CNNs, vision transformer $(\vit)$ models \cite{osovitskiy2021image,vaswani2017attention} can capture long-range visual dependencies due to their flexible receptive fields via self-attention mechanisms. More recently, self-supervised $\vit$s method \cite{wang2021transpath,li2021efficient} combined the advantages of $\vit$ and $\ssl$ to efficiently learn visual representations from less curated pre-training data. Despite their usefulness, there is relatively little research on the impact of data augmentation design, efficiency and robustness of $\vit$ for histopathological image classification. For example, can we train an efficient transformer by selecting only informative regions of high diagnostic value (RoIs) from high-resolution images? What data augmentation strategies can improve the transformer's representation learning for $\trois$ classification? This paper addresses these questions by uncovering insights about key aspects of data augmentation and exploits the self-attention maps to identify the most relevant regions for the end task and train an efficient transformer.   

\vspace{1ex} \noindent \textbf{Contributions.}
\label{par:contribution}
Our contributions are as follows:

\begin{enumerate}
    \item We propose $\scorenet$, a new efficient transformer-based architecture for histopathological image classification. It combines a fine-grained local attention mechanism with a coarse-grained global attention module to extract cell- and tissue-level features. Benefiting from a differentiable recommendation module, the proposed architecture only processes the most discriminative regions of the high-resolution image, making it significantly more efficient than competitive transformer architectures without compromising accuracy;
    
    \item A novel mixing data-augmentation, namely $\scoremix$ for histopathological images is presented. $\scoremix$ works in synergy with our architecture, as they build upon the same observation: the different regions of the images are not equally relevant for a given task. Using the learned self-attention w.r.t. the \texttt{[CLS]} token, we determine the distribution of the semantic regions in images during training to ensure sampling of informed cutting and pasting locations (see Fig.~\ref{fig:scoreMixV2});
    
    
    \item We empirically show consistent improvements of $\scorenet$ over SOTA methods for $\troi$ classification on the BRACS dataset while we demonstrate $\scorenet$'s generalization capability on the CAMELYON16 and BACH datasets. The interpretability of $\scorenet$ behaviour is also investigated. Finally, we demonstrate $\scorenet$ throughput improvements over existing efficient transformers, making it an ideal candidate for applications on $\wsis$.
\end{enumerate}

\begin{figure*}
    \centering
    \includegraphics[width=\textwidth]{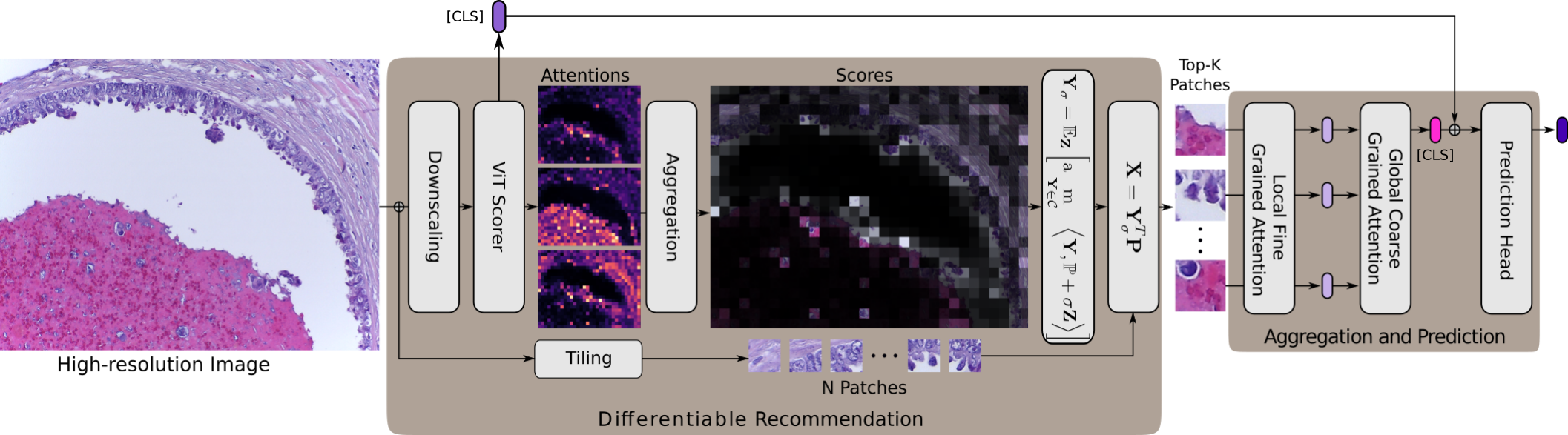}
    \caption{\textbf{An overview of the proposed $\snet$}. The recommendation stage provides tissue-level features, and \textbf{differentiably selects} the most discriminative high-resolution patches. The aggregation stage independently extracts cell-level features and embeds the patches via a \textit{local fine-grained attention} mechanism and endows them with contextual information with the \textit{global coarse-grained attention} mechanism.}
    \label{fig:troi_architecture}
\end{figure*}

\section{Related work}
\label{sec:related}
\vspace{1ex} \noindent \textbf{TRoIs Classification.}
\label{par:trois_classification}
Conventionally, deep convolutional neural networks ~\cite{Srinidhi2021deep,sirinukunwattana2018improving,mercan2019patch,hou2016patch,xu2015deep} process pathology images in a patch-wise manner using a $\mil$ formulation~\cite{maron1998framework} and aggregate patch-level features extracted by $\cnn$s. Nonetheless, current $\mil$ methods discard the inter-patches interaction or only integrate it at the very end of the pipeline. Similarly, the computational resources dedicated to a specific region are independent of its pertinence for the task. Current methods rely on attention-based $\mil$ techniques \cite{ilse2018attention,li2021dual,kalra2021pay,chen2021transmix,rymarczyk2021kernel} to account for the non-uniform relevance of patches. On the contrary, the integration of contextual cues remains almost untouched, as all the aforementioned methods rely on a pipeline where the patch embedding and patch contextualization tasks are disconnected w.r.t. the gradient flow. For example, \cite{kalra2021pay} processes representative patches extracted by an external tool \cite{kalra2020yottixel}. Thus, their patch extraction is fixed and not data-driven as ours. Alternatively, \cite{tokunaga2019adaptive} resort to using a multiple field-of-views/resolutions strategy to endow local patches with contextual information. In another line of research, graph neural network (GNN)-based methods \cite{zhou2019cgc,PATI2022102264} have been proposed to capture global contextual information. These approaches build a graph model that operates on the cell-level structure or combines the cell-level and tissue-level context. However, graph generation can be cumbersome and computationally intensive, prohibiting its use in real-time applications. Recently, SSL methods \cite{li2021dual,koohbanani2021self,srinidhi2021self} have demonstrated their capabilities to improve classification for histopathological images. Most of these methods harness pretext tasks, e.g., contrastive pre-training, to learn semantically meaningful features. Nonetheless, the $\cnn$ backbone used in these approaches inevitably abandons learning of global context features. The transformer-based architectures \cite{wang2021transpath,li2021efficient} can be an alternative solution for processing images as a de-structured patch sequence and capturing their global dependencies. More recently, hybrid-based vision transformer models \cite{chen2021multimodal,shao2021transmil,wang2021transpath} have been used in digital pathology, either based on MIL framework \cite{shao2021transmil} or SSL pre-training \cite{wang2021transpath} on unlabeled histopathological images. Nevertheless, these methods process the whole image uniformly and do not allow dynamic extraction of the region of interest.

\vspace{1ex} \noindent \textbf{Mixing Data-Augmentation Methods.}
Recently, mixing data augmentations strategies \cite{walawalkar2020attentive,yun2019cutmix,yun2019cutmix} have been proposed to enhance the generalization capabilities of deep network classifiers. These improvements are further exacerbated when the augmentations model the interactions between the classes \cite{yun2019cutmix}. These methods create a new augmented sample by cutting an image region from one image and pasting it on another image, while a convex combination of their labels gives the ground-truth label of the new sample. Despite the strong performances of the existing methods, none of them is genuinely satisfying as they either create samples that exhibit atypical local features as in MixUp \cite{zhang2018mixup} or produce potentially mislabeled samples as in CutMix \cite{yun2019cutmix}. CutMix approach has been improved by \cite{chen2021transmix} via re-weighting the mixing factor w.r.t. the sum of the attention map values in the randomly sampled image region, which is still at risk of producing mislabelled samples. In addition, recent CutMix based augmentation methods \cite{walawalkar2020attentive,uddin2020saliencymix} bear additional disadvantages. For example, Attentive CutMix \cite{walawalkar2020attentive} requires an auxiliary pre-trained model to select the most salient patches from the source image and disregards the location of the informative regions in the target image. SaliencyMix \cite{uddin2020saliencymix} assumes that discriminative parts in an image are highly correlated with the saliency map, which is typically not the case for histopathological images.

\section{Methods}
\label{sec:methodo}
\vspace{1ex} \noindent \textbf{Model Overview.} \label{subsec:method:TRoIs_model} An overview of the proposed training pipeline for $\he$ stained histology $\trois$' representation learning is illustrated in Fig.~\ref{fig:troi_architecture}. Histopathological image classification requires capturing cellular and tissue-level microenvironments and learning their respective interactions. Motivated by the above, we propose an efficient transformer, $\scorenet$ that captures the cell-level structure and tissue-level context at the most appropriate resolutions. Provided sufficient contextual information, we postulate and empirically verify that a tissue's identification can be achieved by only attending to its sub-region in a high-resolution image. As a consequence, $\scorenet$ encompasses two stages. The former (\textit{differentiable recommendation}) provides contextual information and selects the most informative high-resolution regions. The latter (\textit{aggregation and prediction}) processes the recommended regions and the global information to identify the tissue and model their interactions simultaneously.

More precisely, the recommendation stage is implemented by a $\vit$ and takes as input a downscaled image to produce a semantic distribution over the high-resolution patches. Then, the most discriminative high-resolution patches for the end task are \textbf{differentiably extracted}. These selected patches (tokens) are then fed to a second $\vit$ implementing the \textit{local fine-grained attention} module, which identifies the tissues represented in each patch. Subsequently, the \textbf{embedded patches attend to one another via a transformer encoder} (\textit{global coarse grained  attention}). This step concurrently refines the tissues' representations and model their interactions. As a final step, the concatenation of the \texttt{[CLS]} tokens from the recommendation's stage and that of the \textit{global coarse-grained attention}'s encoder produces the image's representation. Not only does $\scorenet$'s workflow allows for a significantly increased throughput compared to SOTA methods (see Table~\ref{table:throughput}), it further enables the independent pre-training and validation of its constituent parts.

\subsection{ScoreNet}
\label{subsec:method:scorenet}
\vspace{1ex} \noindent \textbf{Semantic Regions Recommendation.}
\label{subsubsec:patch_selection}
Current $\mil$-based approaches~\cite{ilse2018attention,li2021dual} based on patch-level features aggregation often process histopathological images uniformly and discard the inter-patches interactions. To alleviate these issues, we exploit \textbf{a differentiable recommendation stage to extract discriminative image regions relevant to the classification}. More specifically, we leverage the self-attention map of a $\vit$ as a distribution of the semantic content. Towards that end, the high-resolution image is first downscaled by a factor $s$ and subsequently fed to the recommendation's stage $\vit$. The resulting self-attention map captures the contribution of each patch to the overall representation. Let's assume a $\vit$, that processes a low-resolution image $x_{l} \in \mathbb{R}^{C \times h \times w}$ encompassing $N$ patches of dimension $P_{l} \times P_{l}$. The attended patches (tokens) of the $(L-1)$ layer are conveniently represented as a matrix $\textbf{Z} \in \mathbb{R}^{\left(N+1\right) \times d}$, where $d$ is the embedding dimension of the model, and the extra index is due to the \texttt{[CLS]} token. Up to the last MLP and for a single attention head, the representation of the complete image is given by:
\begin{equation}
\label{eq:y_cls}
    y_{[\text{CLS}]} = \underbrace{\text{softmax}(a_{1}^{T})}_{1 \times (N+1)}\underbrace{\textbf{ZW}_{\text{val}}}_{(N+1) \times d}
\end{equation}
where $\textbf{W}_{\text{val}} \in \mathbb{R}^{d \times d}$ is the value matrix, and $a_{1}^{T}$ is the first row of the self-attention matrix \textbf{A}:
\begin{equation}
    \textbf{A} = \textbf{ZW}_{\text{qry}} \left( \textbf{ZW}_{\text{key}}\right)^T
\end{equation}
where $\textbf{W}_{\text{qry}}$ and $\textbf{W}_{\text{key}}$ are the query and key matrices, respectively. The first row of the self-attention matrix captures the contribution of each token to the overall representation (Eq.~\ref{eq:y_cls}). This is in line with the discriminative capacity of the \texttt{[CLS]} token that patches having the highest contribution are the ones situated in the highest semantic regions of the images. The distribution of the semantic content over the patches is therefore defined as:
\begin{equation}
\label{eq:dist_patch}
    \mathbb{P}_{\text{patch}} = \text{Softmax}(\Tilde{a}_{1}^{T}) \in \mathbb{R}^{N}
\end{equation}
where $\Tilde{a}_{1}$ stands for $a_{1}$ without the first entry, namely the one corresponding to the \texttt{[CLS]} token. Since $\vit$s typically encompasses multiple heads, we propose to add an extra learnable parameter, which weights the relative contributions of each head to the end task; after aggregation of the multiples self-attention maps, the formulation is identical to that of Eq.~\ref{eq:dist_patch}.

Concurrently with acquiring the above defined semantic distribution, the high-resolution image, $x_{h} \in \mathbb{R}^{C \times H \times W}$, is tiled in a regular grid of large patches ($P_{h} \times P_{h}$), stored in a tensor $\textbf{P} \in \mathbb{R}^{N \times C \times P_{h} \times P_{h}}$. At inference time, a convenient way to select the $K$ most semantically relevant high-resolution regions is to encode the \textit{top-K} indices as one-hot vectors: $\textbf{Y} \in \mathbb{R}^{N \times K}$, and to extract the corresponding $K$ patches, $\textbf{X} \in \mathbb{R}^{K \times C \times P_{h} \times P_{h}}$ via: 
\begin{equation}
    \label{eq:patch_extraction}
    \textbf{X} = \textbf{Y}^{T} \textbf{P}
\end{equation}
At training time, since the above formulation is not differentiable, we propose to adopt the differentiable approach of~\cite{cordonnier2021differentiable}. Following the perturbed optimizers scheme, the \textit{top-K} operation is bootstrapped by applying a Gaussian noise, $\sigma \textbf{Z}$, to the semantic distribution. The noisy indicators, $\textbf{Y}_{\sigma}$, are subsequently computed as:
\begin{equation}
    \label{eq:topk_forward}
    \textbf{Y}_{\sigma} = \mathbb{E}_{\mathbf{Z}}\left[ \argmax_{\textbf{Y} \in \mathcal{C} } \left<\textbf{Y}, \Tilde{\mathbb{P}} + \sigma \textbf{Z} \right> \right]
\end{equation}
where $\sigma$ is the standard deviation of the noise, $\Tilde{\mathbb{P}} \in \mathbb{R}^{N \times K}$ is obtained by broadcasting $\mathbb{P}_{\text{patch}}$ to match the dimension of $\textbf{Y}$, and $\mathcal{C}$ is a restriction of the domain ensuring the equivalence between solving Eq.~\ref{eq:topk_forward} and the \textit{top-K} operation \cite{cordonnier2021differentiable}. The extraction of the high-resolution regions follows the procedure described in Eq.~\ref{eq:patch_extraction}. Similarly, the gradient of the indicators w.r.t. the semantic distribution, $\mathbb{P}_{\text{patch}}$ can be computed as:
\begin{equation}
    \label{eq:topk_backward}
        \nabla \textbf{Y}_{\sigma} = \mathbb{E}_{\mathbf{Z}}\left[ \argmax_{\textbf{Y} \in \mathcal{C} } \left<\textbf{Y}, \Tilde{\mathbb{P}} + \sigma \textbf{Z} \right> \textbf{Z}^{T} / \sigma \right]
\end{equation}

\vspace{1ex} \noindent \textbf{Computational Complexity.}
\label{subsubsec:comp_comp}
Vision transformers heavily rely on the attention mechanism to learn a high-level representation from low-level regions. The underlying assumption is that the different sub-regions of the image are not equally important for the overall representation. Despite this key observation, the computation cost dedicated to a sub-region is independent of its contribution to the high-level representation, which is inefficient. Our $\scorenet$ attention mechanism overcomes this drawback by learning to attribute more resources to regions of high interest. For a high-resolution input image $x_{h} \in \mathbb{R}^{C \times H \times W}$, the asymptotical time and memory cost is $\mathcal{O} \left(\left( \frac{H}{s \cdot P_{l}} \cdot \frac{W}{s \cdot P_{l}}\right)^2\right)$, when the recommendation stage uses inputs downscaled by a factor $s$ and processes them with a patch size of $P_{l}$. The derivation of this cost, including that of the recommendation stage, which is independent of the input size, can be found in the \textbf{Appendix}.

\subsection{ScoreMix}
\label{subsec:method:score_mix}
We propose a new mixing data augmentation for histopathological images by learning the \textbf{distribution of the semantic image regions} using the learned self-attention for \texttt{[CLS]} token of the $\vit$ without requiring architectural changes or additional loss. 
\begin{figure}[!t]
    \centering
    \includegraphics[width=\columnwidth]{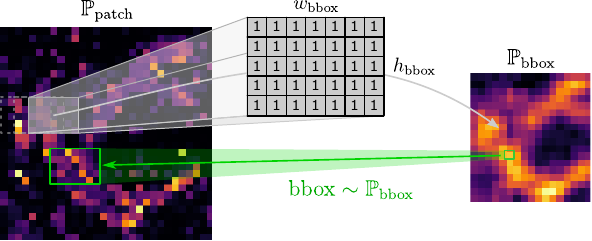}
    \caption{\textbf{The bounding box selection scheme for ScoreMix.} The score distribution for each bounding box ($\mathbb{P}_{\text{bbox}}$) is obtained by convolving the patch distribution map $\mathbb{P}_{\text{patch}}$ with a kernel of $\textbf{1}$s of the bounding box dimensions $(h_{\text{bbox}}, w_{\text{bbox}})$. The bbox is then sampled from $\mathbb{P}_{\text{bbox}}$, which we often refer to as $\mathbb{P}_{\text{source}}$ or $\mathbb{P}_{\text{target}}$.}
    \label{fig:bbox_sampling}
\end{figure}
More formally, let $x_{s}, x_{t} \in \mathbb{R}^{C \times H \times W}$ be the source and target images respectively and let $y_{s}$ and $y_{t}$ be their corresponding labels. We aim to mix the source and target samples to generate a new training example $(x_{m}, y_{m})$. To do so, we first compute the semantic distributions using the current parameters of the model and the input samples; namely, we compute $\mathbb{P}_{\text{source}}(x_{s}, \theta)$ and $\mathbb{P}_{\text{target}}(x_{t}, \theta)$. Given these distributions and a randomly defined bounding box size, we sample the cutting and pasting locations from the source and target distributions, respectively:
\begin{equation}
    \begin{aligned}
    M_{s} & \sim \frac{1}{Z_{s}} \cdot \mathbb{P}_{\text{source}}(x_{s}, \theta, \lambda) \\
    M_{t} & \sim \frac{1}{Z_{t}} \cdot \left(1 - \mathbb{P}_{\text{target}}(x_{t}, \theta, \lambda)\right) \end{aligned}
\end{equation}
where $Z_{s}$ and $Z_{t}$ are normalization constants, and $1-\lambda \sim \mathcal{U}([0, 1])$ defines the strength of the mixing, i.e. the size of the bounding box. The locations of the cutting and pasting regions are encoded as binary masks, i.e., $M_{s}, M_{t} \in \{0, 1\}^{H \times W}$, where a value of $1$ encodes for a patch in the cutting/pasting region. Under the above formalism, the mixing operation can be defined as:
\begin{equation}
\begin{aligned}
        x_{m} = (\textbf{1} - M_{t}) \odot x_{t} \\
        M_{t} \otimes x_{m} \leftarrow M_{s} \otimes x_{s} \\
        y_{m} = \lambda y_{t} + (1 - \lambda) y_{s}
\end{aligned}
\end{equation}
where $\textbf{1}$ is a mask of ones, $\odot$ denotes the element-wise multiplication, and $\otimes$ indicates an indexing w.r.t. a mask.
\vspace{1ex} \noindent \textbf{Computing the Semantic Distributions.}
Computing the semantic distributions of the target and source images is an essential part of the pipeline as it allows for a data-driven selection of the cutting/pasting sites, thereby avoiding the pitfalls of random selection. When the size of the bounding box matches that of a single patch, the distribution can be directly deduced from the self-attention map, as described in Sec.~\ref{subsec:method:TRoIs_model}.
As a consequence, and when the bounding box's size matches that of a single patch, the semantic distribution can be directly obtained from $\mathbb{P}_{\text{patch}}$ (see Eq.~\ref{eq:dist_patch}).
In practice, we would typically use bounding boxes encompassing more than a single patch. In that case, the distribution of the semantic content at the bounding box resolution can be obtained by a local aggregation of the distribution above:
\begin{equation}
\mathbb{P}_{\text{bbox}}(i) \propto \sum\nolimits_{j \in \mathcal{N}(i)}\mathbb{P}_{\text{patch}}(j)
\end{equation}
where $\mathcal{N}(i)$ returns the indices of the patches situated in the bounding box whose top left corner is the patch $i$. In practice, this can be efficiently implemented by first unflattening the patch distribution $\mathbb{P}_{\text{patch}}$, and convolving it with a kernel of ones and of the same dimension as the desired bounding box (see Fig.~\ref{fig:bbox_sampling}).
\section{Experiments}
\label{sec:experiments}
\vspace{1ex} \noindent \textbf{Datasets.}
The primary dataset used in our experiments is the BReAst Carcinoma Sub-typing (\textbf{BRACS})~\cite{PATI2022102264}. BRACS consists of $4391$ RoIs acquired from $325$ $\he$ stained breast carcinoma $\wsi$ (at $0.25$ $\mu$m/pixel) with varying dimensions and appearances. Each RoI is annotated with one of the seven classes: Normal, Benign, Usual Ductal Hyperplasia (UDH), Atypical Ductal Hyperplasia (ADH), Flat Epithelial Atypia (FEA), Ductal Carcinoma In Situ (DCIS), and Invasive. Our experiments follow the same data splitting scheme as \cite{PATI2022102264} for training, validation, and test set at the $\wsi$ level to avoid test leakage. In addition, we use publicly available BreAst Cancer Histology \textbf{(BACH)} dataset~\cite{ARESTA2019122} to show $\scorenet$ generalization capabilities. It contains $400$ training and $100$ test images from four different breast cancer types: Normal,  Benign, In Situ, and Invasive. All images have a fixed size of $1536 \times 2048$ pixels and a pixel scale of $0.42 \times 0.42$ $\mu$m. To assess the interpretability of $\scorenet$, we further evaluate our model on the \textbf{CAMELYON16} dataset~\cite{camelyon16} for binary tumour classification. We extract a class-balanced dataset of $1920\times1920$ pixels from high-resolution $\wsi$s.

\begin{table*}[ht]
\centering
\caption{\textbf{Comparison with the prior art for $\TR$ classification} using weighted and class-wise F1-scores averaged over three independent runs on the BRACS dataset. The best results are in \textbf{bold}. $\mathrm{ScoreNet/x/y}$ refers to an instance of $\scorenet$ using the recommendation module's last x [CLS] tokens and the last y tokens from the global coarse-grained attention.}
\resizebox{\textwidth}{!}{
\begin{tabular}{c | l c c c c c c c | c}
\toprule 
& Method         & Normal & Benign & UDH & ADH & FEA & DCIS & Invasive & Weighted F1  \\
\midrule 
\multirow{17}{0.3cm}{\centering \rot{$\mil$s}} & Agg-Penultimate ($10\times$) \cite{sirinukunwattana2018improving} & 48.7 $\pm$ 1.7 & 44.3 $\pm$ 1.9 & 45.0 $\pm$ 5.0 & 24.0 $\pm$ 2.8 & 47.0 $\pm$ 4.3  & 53.3 $\pm$ 2.6 & 86.7 $\pm$ 2.6 & 50.8 $\pm$ 2.6 \\
 & Agg-Penultimate ($20\times$) \cite{sirinukunwattana2018improving}    & 42.0 $\pm$ 2.2 & 42.3 $\pm$ 3.1 & 39.3 $\pm$ 2.0 & 22.7 $\pm$ 2.5 & 47.7 $\pm$ 1.2  & 50.3 $\pm$ 3.1 & 77.0 $\pm$ 1.4 & 46.8 $\pm$ 2.2 \\
 & Agg-Penultimate  ($40\times$) \cite{sirinukunwattana2018improving}  & 32.3 $\pm$ 4.6 & 39.0 $\pm$ 0.8 & 23.7 $\pm$ 1.7 & 18.0 $\pm$ 0.8 & 37.7 $\pm$ 2.9  & 47.3 $\pm$ 2.0 & 70.7 $\pm$ 0.5 & 39.4 $\pm$ 1.9 \\
 & Agg-Penultimate ($10\times$ + $20\times$) \cite{sirinukunwattana2018improving}  & 48.3 $\pm$ 2.0 & 45.7 $\pm$ 0.5 & 41.7 $\pm$ 5.0 & 32.3 $\pm$ 0.9 & 46.3 $\pm$ 1.4  & 59.3 $\pm$ 2.0 & 85.7 $\pm$ 1.9 & 52.3 $\pm$ 1.9 \\
 & Agg-Penultimate  ($10\times$ + $20\times$ + $40\times$) \cite{sirinukunwattana2018improving}  & 50.3 $\pm$ 0.9 & 44.3 $\pm$ 1.2 & 41.3 $\pm$ 2.5 & 31.7 $\pm$ 3.3 & 51.7 $\pm$ 3.1  & 57.3 $\pm$ 0.9 & 86.0 $\pm$ 1.4 & 52.8 $\pm$ 1.9 \\
 & CLAM-SB/S ($10\times$) \cite{lu2021data} & 39.6 $\pm$ 4.6 &  45.5 $\pm$ 4.9 & 34.7 $\pm$ 2.0 & 30.4 $\pm$ 6.7 & 68.8 $\pm$ 1.9 & 64.3 $\pm$ 0.8 & 84.2 $\pm$ 2.6 & 53.9 $\pm$ 1.9 \\
 & CLAM-SB/S ($20\times$) \cite{lu2021data} & 50.2 $\pm$ 3.2 & 45.5 $\pm$ 1.8 & 32.2 $\pm$ 1.6 & 25.5 $\pm$ 4.2 & 69.6 $\pm$ 1.0 & 60.8 $\pm$ 2.7 & 84.2 $\pm$ 1.6 & 54.0 $\pm$ 0.7 \\
 & CLAM-SB/S ($40\times$) \cite{lu2021data} & 47.0 $\pm$ 5.2 & 38.8 $\pm$ 1.8 & 30.0 $\pm$ 7.7 & 29.4 $\pm$ 2.9 & 65.9 $\pm$ 1.2 & 52.2 $\pm$ 1.3 & 76.7 $\pm$ 1.6 & 49.9 $\pm$ 0.8 \\
 & CLAM-SB/B ($10\times$) \cite{lu2021data} & 46.4 $\pm$ 6.0 & 42.4 $\pm$ 2.8 & 33.1 $\pm$ 1.0 & 29.3 $\pm$ 2.1 & 67.4 $\pm$ 1.4 & 63.0 $\pm$ 4.5 & 84.4 $\pm$ 2.1 & 53.7 $\pm$ 1.9 \\
 & CLAM-SB/B ($20\times$) \cite{lu2021data} & 56.2 $\pm$ 1.2 & 42.3 $\pm$ 4.4 & 27.4 $\pm$ 2.4 & 30.1 $\pm$ 4.0 & 68.5 $\pm$ 2.1 & 60.9 $\pm$ 2.1 & 84.6 $\pm$ 1.2 & 54.3 $\pm$ 1.5 \\
 & CLAM-SB/B ($40\times$) \cite{lu2021data} & 42.8 $\pm$ 1.1 & 43.3 $\pm$ 2.8 & 33.8 $\pm$ 0.7 & 29.6 $\pm$ 3.6 & 64.1 $\pm$ 2.6 & 52.0 $\pm$ 3.8 & 78.8 $\pm$ 2.2 & 50.5 $\pm$ 0.9 \\
 & CLAM-MB/S ($10\times$) \cite{lu2021data}& 42.5 $\pm$ 3.3 & 43.4 $\pm$ 3.6 & 31.4 $\pm$ 3.2 & 32.1 $\pm$ 4.8 & 67.5 $\pm$ 2.2 & 59.7 $\pm$ 2.4 & 83.8 $\pm$ 2.0 & 52.9 $\pm$ 1.7 \\
 & CLAM-MB/S ($20\times$) \cite{lu2021data}& 56.6 $\pm$ 0.8 & 47.4 $\pm$ 0.9 & 33.5 $\pm$ 5.2 & 17.0 $\pm$ 1.5 & 70.3 $\pm$ 1.1 & 56.9 $\pm$ 1.6 & 84.9 $\pm$ 1.2 & 53.8 $\pm$ 0.6 \\
 & CLAM-MB/S ($40\times$) \cite{lu2021data}& 50.2 $\pm$ 7.7 & 39.3 $\pm$ 2.9 & 38.6 $\pm$ 2.4 & 26.5 $\pm$ 8.9 & 69.4 $\pm$ 2.6 & 54.1 $\pm$ 3.3 & 82.9 $\pm$ 2.5 & 52.9 $\pm$ 0.8 \\
 & CLAM-MB/B ($10\times$) \cite{lu2021data}& 39.7 $\pm$ 1.6 & 41.0 $\pm$ 2.6 & 34.5 $\pm$ 1.0 & 29.8 $\pm$ 4.7 & 66.8 $\pm$ 1.5 & 63.4 $\pm$ 1.0 & 83.5 $\pm$ 0.4 & 52.7 $\pm$ 0.9 \\
 & CLAM-MB/B ($20\times$) \cite{lu2021data}& 59.4 $\pm$ 2.0 & 47.7 $\pm$ 1.2 & 31.7 $\pm$ 0.7 & 20.1 $\pm$ 3.4 & 68.3 $\pm$ 0.4 & 59.9 $\pm$ 1.7 & 86.8 $\pm$ 0.6 & 54.8 $\pm$ 1.0 \\
 & CLAM-MB/B ($40\times$) \cite{lu2021data}& 47.3 $\pm$ 3.2 & 39.5 $\pm$ 1.5 & 38.8 $\pm$ 4.5 & 30.2 $\pm$ 6.3 & 68.2 $\pm$ 1.9 & 59.2 $\pm$ 2.9 & 82.1 $\pm$ 2.7 & 53.5 $\pm$ 1.3 \\
 \midrule 
\multirow{8}{0.3cm}{\centering \rot{GNNs}} & CGC-Net \cite{zhou2019cgc}  & 30.8 $\pm$ 5.3 & 31.6 $\pm$ 4.7 & 17.3 $\pm$ 3.4 & 24.5 $\pm$ 5.2 & 59.0 $\pm$ 3.6  & 49.4 $\pm$ 3.4 & 75.3 $\pm$ 3.2 & 43.6 $\pm$ 0.5 \\
 & Patch-GNN ($10\times$) \cite{aygunecs2020graph}  & 52.5 $\pm$ 3.3 & 47.6 $\pm$ 2.2 & 23.7 $\pm$ 4.6 & 30.7 $\pm$ 1.8 & 60.7 $\pm$ 5.3  & 58.8 $\pm$ 1.1 & 81.6 $\pm$ 2.2 & 52.1 $\pm$ 0.6 \\
 & Patch-GNN ($20\times$) \cite{aygunecs2020graph}  & 43.9 $\pm$ 4.2 & 43.4 $\pm$ 3.2 & 19.5 $\pm$ 2.3 & 25.7 $\pm$ 2.9 & 55.6 $\pm$ 2.1  & 52.9 $\pm$ 1.8 & 79.2 $\pm$ 1.1 & 47.1 $\pm$ 0.7 \\
 & Patch-GNN ($40\times$) \cite{aygunecs2020graph}  & 41.7 $\pm$ 3.1 & 32.9 $\pm$ 1.0 & 25.1 $\pm$ 3.7 & 25.6 $\pm$ 2.0 & 49.5 $\pm$ 3.5  & 48.6 $\pm$ 4.2 & 71.6 $\pm$ 5.1 & 43.2 $\pm$ 0.6 \\
 & TG-GNN \cite{pati2020hact}  & 58.8 $\pm$ 6.8 & 40.9 $\pm$ 3.0 & 46.8 $\pm$ 1.9 & 40.0 $\pm$ 3.6 & 63.7 $\pm$ 10.5 & 53.8 $\pm$ 3.9 & 81.1 $\pm$ 3.3 & 55.9 $\pm$ 1.0 \\
 & CG-GNN \cite{pati2020hact}   & 63.6 $\pm$ 4.9 & 47.7 $\pm$ 2.9 & 39.4 $\pm$ 4.7 & 28.5 $\pm$ 4.3 & 72.1 $\pm$ 1.3  & 54.6 $\pm$ 2.2 & 82.2 $\pm$ 4.0 & 56.6 $\pm$ 1.3 \\
 & CONCAT-GNN   & 61.0 $\pm$ 4.5 & 43.1 $\pm$ 2.3 & 42.0 $\pm$ 4.7 & 26.1 $\pm$ 3.7 & 71.3 $\pm$ 2.1  & 60.8 $\pm$ 3.7 & 85.4 $\pm$ 2.7 & 57.0 $\pm$ 2.3 \\
 & HACT-Net \cite{pati2020hact} & 61.6 $\pm$ 2.1 & 47.5 $\pm$ 2.9 & 43.6 $\pm$ 1.9 & 40.4 $\pm$ 2.5 & 74.2 $\pm$ 1.4  & \textbf{66.4 $\pm$ 2.6} & 88.4 $\pm$ 0.2 & 61.5 $\pm$ 0.9\\
 \midrule 
 \multirow{8}{0.3cm}{\centering \rot{Tansformers}} & TransPath \cite{wang2021transpath}  & 58.5 $\pm$ 2.5 & 43.1 $\pm$ 1.8 & 34.9 $\pm$ 5.2 & 38.3 $\pm$	6.0 & 66.9 $\pm$ 0.8 & 61.4 $\pm$ 1.2 & 85.0 $\pm$1.4 & 56.7 $\pm$ 2.0\\
 & TransMIL ($10\times$) \cite{shao2021transmil} & 38.7 $\pm$ 5.4 & 44.0 $\pm$ 2.9 & 30.5 $\pm$ 4.1 & 31.0 $\pm$ 11.8 &  68.1 $\pm$ 2.6 & 61.8 $\pm$ 1.9 & 87.3 $\pm$ 2.6 & 53.2 $\pm$ 1.1 \\
 & TransMIL ($20\times$) \cite{shao2021transmil} & 51.0 $\pm$ 0.1 & 44.5 $\pm$ 2.9 & 31.6 $\pm$ 2.1 & 31.4 $\pm$ 10.3 & 71.3 $\pm$ 4.8 & 63.0 $\pm$ 2.8 & 89.9 $\pm$ 1.6 & 56.2 $\pm$ 1.6 \\
 & TransMIL ($40\times$) \cite{shao2021transmil} & 47.6 $\pm$ 9.8 & 42.9 $\pm$ 3.6 & 41.5 $\pm$ 5.3 & 38.4 $\pm$ 5.9 & 72.7 $\pm$ 2.6 & 62.7 $\pm$ 2.9 & 87.1 $\pm$ 3.9 & 57.5 $\pm$ 0.7 \\
 \cmidrule(l){2-10}
 & \cellcolor{violet!20} Lin. encoder's [CLS] & \cellcolor{violet!20} 52.7 $\pm$ 9.4 & \cellcolor{violet!20} 35.6 $\pm$ 3.4 & \cellcolor{violet!20} 34.5 $\pm$ 6.7 & \cellcolor{violet!20} 25.1 $\pm$ 3.6 & \cellcolor{violet!20} 53.5 $\pm$ 9.8 & \cellcolor{violet!20} 38.7 $\pm$ 2.8 & \cellcolor{violet!20} 63.3 $\pm$ 7.6 & \cellcolor{violet!20} 43.8 $\pm$ 3.4 \\
 & \cellcolor{violet!20} Lin. scorer's [CLS] & \cellcolor{violet!20} 57.5 $\pm$ 4.2 & \cellcolor{violet!20} 48.8 $\pm$ 5.5 & \cellcolor{violet!20} 42.7 $\pm$ 3.5 & \cellcolor{violet!20} 42.7 $\pm$ 7.4 & \cellcolor{violet!20} 74.3 $\pm$ 5.2 & \cellcolor{violet!20} 60.5 $\pm$ 2.4 & \cellcolor{violet!20} 90.6 $\pm$ 0.2 & \cellcolor{violet!20} 60.9 $\pm$ 3.1 \\
 \cmidrule(l){2-10}
 & \cellcolor{violet!20} $\mathrm{ScoreNet/4/1}$ & \cellcolor{violet!20} \textbf{64.6 $\pm$ 2.2} & \cellcolor{violet!20} 52.6 $\pm$ 2.8 & \cellcolor{violet!20} \textbf{48.4 $\pm$ 2.2} & \cellcolor{violet!20} \textbf{47.4 $\pm$ 2.4} & \cellcolor{violet!20} 77.9 $\pm$ 0.7 & \cellcolor{violet!20} 59.3 $\pm$ 1.1 & \cellcolor{violet!20} 90.6 $\pm$ 1.5 & \cellcolor{violet!20} 64.1 $\pm$ 0.7 \\
 & \cellcolor{violet!20} $\mathrm{ScoreNet/4/3}$ & \cellcolor{violet!20}64.3 $\pm$ 1.5 & \cellcolor{violet!20} \textbf{54.0 $\pm$ 2.2} & \cellcolor{violet!20} 45.3 $\pm$ 3.4 & \cellcolor{violet!20} 46.7 $\pm$ 1.0 & \cellcolor{violet!20} \textbf{78.1 $\pm$ 2.8} & \cellcolor{violet!20} 62.9 $\pm$ 2.0 & \cellcolor{violet!20} \textbf{91.0 $\pm$ 1.4} & \cellcolor{violet!20} \textbf{64.4 $\pm$ 0.9} \\
\bottomrule 
\end{tabular}}
\vspace{-1.em}
\label{table:f1_scores_trois}
\end{table*}

\vspace{1ex} \noindent \textbf{Experimental Setup.}
\label{sec:experimentalsetting}
We base $\scorenet$' ViTs, namely the one used by the recommendation stage and by the local fine-grained attention mechanism on a modified ViT-tiny architecture (see \textbf{Appendix}) and follow the self-supervised pre-training scheme of \cite{caron2021emerging} for both of the aforementioned $\vit$s. Noteworthy that an end-to-end pre-training of $\scorenet$ is also feasible.
After pre-training, the $\scorenet$ is optimized using the SGD optimizer (momentum=0.9) with a learning rate chosen with the linear scaling rule \cite{goyal2018accurate} ($lr=10^{-2}\cdot \text{batchsize} / 256 = 3.125\cdot10^{-4}$) annealed with a cosine schedule until $10^{-6}$. $\scorenet$ is finetuned for $15$ epochs with a batch-size of $8$. We empirically determine the top $K=20$ regions, and a downscaling factor $s=8$ by a hyperparameter sweep (cf. ablation experiment in the \textbf{Appendix}).
All experiments are implemented in PyTorch 1.9 \cite{paszke2019pytorch} using a single GeForce RTX3070 GPU.

\subsection{TRoIs Classification Results and Discussion}
\label{subsubsec:trois_classification}
In Table~\ref{table:f1_scores_trois}, we compare the $\trois$ classification performance of $\scorenet$ on the BRACS dataset against the state-of-the-arts, including MIL-based~\cite{mercan2019patch,shao2021transmil,lu2021data}, GNN-based, e.g.,~\cite{PATI2022102264}, and self-supervised transformer-based~\cite{wang2021transpath} approaches. The first $\mil$-based baseline~\cite{mercan2019patch} aggregates independent patch representations from the penultimate layer of a ResNet-50 \cite{he2016deep} pre-trained on ImageNet \cite{5206848}. The patch model is further finetuned on $128\times128$ patches at different magnification, e.g., $10\times$, $20\times$ or $40\times$. The latter operate either on multi- or single-scale images to benefit from varying levels of context and resolution. Similarly, we report the performances of the recent MIL-based methods, TransMIL \cite{shao2021transmil}, and CLAM \cite{lu2021data} using the original implementations and setup. Both methods are tested with different magnifications (see Table \ref{table:f1_scores_trois}). Additionally, the single-head (-SB) and multi-head (-MB) variants of CLAM are used with the small (-S) and big (-B) versions of the models (see CLAM's implementation). We further use various GNN-based baselines, particularly HACT-Net \cite{PATI2022102264}, the current SOTA approach for $\troi$ classification on the BRACS. Finally, we report the performance of the recent self-supervised transformer approach, TransPath \cite{wang2021transpath}, which is a hybrid transformer/convolution-based architecture. $\scorenet$ reaches a new state-of-the-art weighted F1-score of $64.4\%$ on the BRACS $\troi$ classification task outperforming the second-best method, HACT-Net, by a margin of $2.9\%$ (Table \ref{table:f1_scores_trois}). The results are reported for two variants of  $\scorenet$, namely $\mathrm{ScoreNet/4/1}$ and $\mathrm{ScoreNet/4/3}$, which use the four last \texttt{[CLS]} tokens of the scorer and the last or the three last \texttt{[CLS]} tokens from the coarse attention mechanism (aggregation stage). $\mathrm{ScoreNet/4/3}$ variant puts more emphasis on the features available at (40$\times$), whereas $\mathrm{ScoreNet/4/1}$ is more biased towards the global representation available at (5$\times$) (with a downscaling factor $s=8$). One can observe that both model variants significantly outperform the existing baseline in terms of weighted F1-scores and for almost every class. More interestingly, the architectural differences directly translate to differences in the classification results. $\mathrm{ScoreNet/4/3}$ is more suitable for classes where the \textbf{discriminative features are at the cell level than $\mathrm{ScoreNet/4/1}$, which is more suited when the tissue organization is the discriminative criterion}. Nonetheless, both of these architectures indeed benefit from the information available at each scale. This observation is well supported by the classification results obtained when a linear layer is trained independently on the scorer's \texttt{[CLS]} tokens (Lin. scorer's \texttt{[CLS]} in Table \ref{table:f1_scores_trois}) or using only the \texttt{[CLS]} tokens from the aggregation stage (Lin. encoder's \texttt{[CLS]} in Table \ref{table:f1_scores_trois}). Despite the difference in results between the two model variants, it is clear that they both perform worse when separated, which indicates that the representations of both stages are complementary. In brief, $\scorenet$ allows for an easily tuning to meet prior inductive biases on the ideal scale for a given task.

\vspace{1ex} \noindent \textbf{ScoreMix \& Data-Regime Sensitivity.}
\label{par:scoremix_results}
We also show that $\scorenet$ equipped with the proposed $\scoremix$ augmentation achieves superior TRoIs classification performances compared to CutMix \cite{yun2019cutmix} and SaliencyMix \cite{uddin2020saliencymix} augmentations for different data regimes, e.g., low-regime with only $10\%$ of the data. \textbf{Our proposed $\score$ outperforms SOTA methods with only 50\% of the data} and is on-par or better than most baselines with only $20\%$ of the data (Table \ref{table:augmentation_comp}). We argue that these improvements are primarily due to the generation of more coherent sample-label pairs under the guidance of the learned semantic distribution. This alleviates randomly cutting and pasting non-discriminative patches, as is the case with CutMix. Our results further support that image saliency used in the SaliencyMix is not correlated with discriminative regions.

\begin{table}[!t]
\centering
\caption{\textbf{Comparison with SOTA Mixup-based augmentation methods \cite{yun2019cutmix,uddin2020saliencymix} and the standard random augmentation strategy} using various fractions of the BRACS dataset and identical distribution for the bounding boxes' sizes.}
\resizebox{\columnwidth}{!}{
\begin{tabular}{l c c c c}
\toprule 
Dataset & Random Aug. & CutMix \cite{yun2019cutmix} & SaliencyMix \cite{uddin2020saliencymix} & \cellcolor{violet!20} $\smix$ \\ 
\midrule 
BRACS 10\%  & 52.9 $\pm$ 2.4 & 53.7 $\pm$ 2.9 & 53.5 $\pm$ 2.7 & \cellcolor{violet!20}\textbf{55.9 $\pm$ 1.9} \\ 
BRACS 20\%  & 57.6 $\pm$ 1.8 & 58.0 $\pm$	1.4 & 57.8 $\pm$ 1.0 & \cellcolor{violet!20}\textbf{58.7 $\pm$ 0.8} \\ 
BRACS 50\%  & 60.4 $\pm$ 1.8 & 61.2 $\pm$ 2.5 & 59.8 $\pm$ 2.4 & \cellcolor{violet!20}\textbf{62.3 $\pm$ 0.6} \\ 
BRACS 100\% & 62.7 $\pm$ 1.6 & 63.1 $\pm$ 1.1 & 62.8 $\pm$ 1.2 & \cellcolor{violet!20}\textbf{64.0 $\pm$ 0.7} \\ 

\bottomrule 
\end{tabular}}
\label{table:augmentation_comp}
\end{table}
\vspace{1ex} \noindent \textbf{Generalization Capabilities.}
\label{subsubsec:generalization_camelyon}
To gauge the generalization capabilities of $\scorenet$ compared to other current SOTA methods, e.g., HACT-Net~\cite{pati2020hact}, we leverage two external evaluation datasets, namely CAMELYON16 and BACH. After training on the BRACS dataset, the weights of $\scorenet$ are frozen. To evaluate the quality of the learned features, we either train a linear classifier on top of the frozen features or apply a $k$-nearest-neighbor classifier ($k$ = 1) without any finetuning. We perform stratified 5-fold cross-validation. For HACT-Net, we use the available pre-trained weights and follow the implementation of~\cite{PATI2022102264}. As HACT-Net sometimes fails to generate embeddings and to have a fair comparison, we only evaluate the samples for which HACT-Net could successfully produce embeddings (around $95\%$ of the BACH and $80\%$ of the CAMELYON16 datasets). Experimental results in Table~\ref{tab:generalization} demonstrate the superiority of $\scorenet$ in learning generalizable features. It further demonstrates the robustness of $\scorenet$ to changes in magnification. Indeed, the model is pre-trained on BRACS (40$\times$), while BACH’s images were acquired at a magnification of 20$\times$. Furthermore, the CAMELYON16 dataset contains $\wsis$ collected from lymph nodes in the vicinity of the breast, while BRACS contains $\wsis$ collected by mastectomy or biopsy (i.e., directly in the breast). The excellent knowledge transfer between the two datasets highlights the transferability of features learned by $\scorenet$ in various use cases.
\begin{table}[!t]
\caption{\textbf{Generalization capabilities} of $\snet$ compared to HACT-Net trained on BRACS and evaluated on the BACH's annotated images and $1000$ images from CAMELYON16, respectively. The weighted F1-scores
over a stratified 5-fold cross-validation fold is reported.}
\resizebox{\columnwidth}{!}{
\centering
\begin{tabular}{l c c c c}
\toprule 
                             & \multicolumn{2}{c}{BRACS $\rightarrow$ BACH} & \multicolumn{2}{c}{BRACS $\rightarrow$ CAMELYON16}    \\ 
                             \cmidrule(lr){2-3}  \cmidrule(lr){4-5}
                             & Linear                    & $k$-NN                & Linear                 & $k$-NN       \\ \midrule 
TransPath \cite{wang2021transpath} & 61.8 $\pm$ 4.8  & 72.0 $\pm$ 2.9 & 58.1 $\pm$ 4.8 & 69.9 $\pm$ 2.5\\ 
TransMIL \cite{shao2021transmil} & 46.5 $\pm$ 10.2 & 74.0 $\pm$ 4.8 & 59.8 $\pm$ 3.0 & 60.8 $\pm$ 5.3 \\ 
CLAM-SB/S \cite{lu2021data} & 53.3 $\pm$ 13.0 & 69.8 $\pm$ 4.5 & 56.7 $\pm$ 1.9 & 68.0 $\pm$ 3.5\\
CLAM-SB/B \cite{lu2021data} & 57.5 $\pm$ 3.6  & 75.3 $\pm$ 3.1 & 55.5 $\pm$ 4.1 & 68.0 $\pm$ 1.5 \\
HACT-Net \cite{pati2020hact}  & 40.2 $\pm$ 2.8  & 32.8 $\pm$ 5.8 & 60.0 $\pm$ 4.6 & 61.0 $\pm$ 4.2 \\ 
\midrule
\cellcolor{violet!20} $\snet$ & \cellcolor{violet!20}\textbf{73.4 $\pm$ 3.5 } & \cellcolor{violet!20}\textbf{76.9 $\pm$ 6.1}    & \cellcolor{violet!20}\textbf{81.1 $\pm$ 3.5  }    & \cellcolor{violet!20}\textbf{77.0 $\pm$ 4.6 }     \\
\bottomrule 
\end{tabular}%
}
\label{tab:generalization}
\end{table}

\begin{figure}[!t]
    \centering
    \includegraphics[width=\columnwidth]{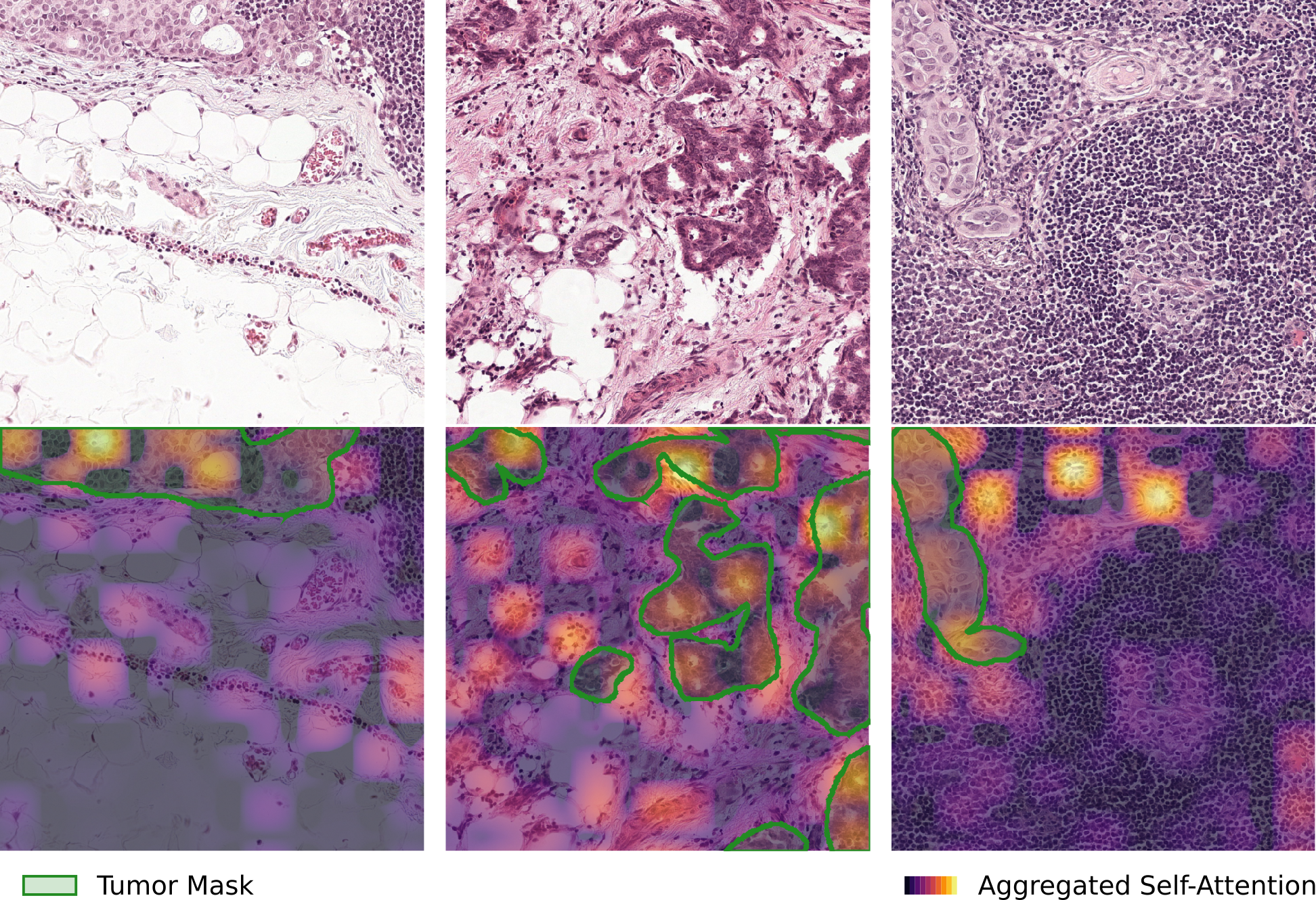}
    \caption{\textbf{$\snet$ Interpretability}. Visualization of the semantic distribution, overlaid with the tumour ground-truth mask on a few samples of the CAMELYON16 dataset. The semantic distributions are obtained from the recommendation stage, i.e., at low-resolution. $\scorenet$ is pre-trained on BRACS and finetuned on CAMELYON16.}
    \label{fig:interpretability}
\end{figure} 

\vspace{1ex} \noindent \textbf{Interpretability?}
\label{subsubsec:interpretability}
To probe the internal behavior of $\scorenet$, we finetune the model on CAMELYON16 images using image-level labels only. At test time, we scrutinize the learned semantic distributions of the tumour-positive images. The semantic distributions depicted in Fig.~\ref{fig:interpretability} seems to indicate that $\scorenet$ learns to identify the tumour area and interpret \textbf{cancer-related morphological information}, while never having been taught to do so.
Quantitatively, we observe that, on average, $74.6 \%$ of the $20$ patches selected from positive images are tumour-positive. Furthermore, we report an average image-wise AuC of 73.6\% when interpreting the probability of the recommendation stage to sample a patch as the probability of it being tumour-positive. 

 \vspace{1ex} \noindent \textbf{Ablation on Efficacy of $\snet$.}
 \label{subsubsec:efficiency}
 The critical aspect of $\scorenet$ is its improved efficiency compared to other transformer-based architectures. This improvement is due to the choice of a hierarchical architecture and the exploitation of redundancy in histology images. At inference time, we expect a gain in throughput compared to the vanilla $\vit$ of the order of the squared downscaling factor, $s$, (see \textbf{Appendix}), typically $s^2 = 64$, which is well reflected in practice, as shown in Table~\ref{table:throughput}. Due to the self-supervised pre-training, $\scorenet$ does not require any stain normalization or pre-processing, unlike its competitor HACT-Net. Similarly, ScoreNet yields higher throughput than other SOTA efficient transformers architectures, namely TransPath \cite{wang2021transpath}, and SwinTransformer \cite{liu2021swin}, with throughput around $3\times$ and $4\times$ higher than these methods. The latter observation is interesting considering the linear asymptotic time and memory cost of the SwinTransformer, which is probably a consequence of the fact that SwinTransformers process a lot of uninformative high-resolution patches in the first layer(s). 
 
\begin{table}[!t]
\centering
\caption{\textbf{Inference throughput comparison of $\snet$, HACT-Net, and SOTA transformer-based architectures}. All models were tested with the same image size and a single GeForce RTX 3070 GPU.}
\resizebox{\columnwidth}{!}{
\begin{tabular}{l c c c}
\toprule 
                     & Image size & Throughput (im./s) & Pre-processing \\
\midrule 
HACT-Net \cite{pati2020hact}           & $1536 \times 2048$ & 4.95$e$-4 $\pm$ 1.40$e$-3 & \ck \\ 
Vanilla ViT \cite{osovitskiy2021image} & $1536 \times 2048$ & 3.8 $\pm$ 0.1 & - \\ 
SwinTransformer \cite{liu2021swin}  & $1536 \times 2048$ & 76.8 $\pm$ 0.4 & \xk \\
TransPath \cite{wang2021transpath} & $1536 \times 2048$ & 97.6 $\pm$ 3.1 & \xk \\
\midrule
\cellcolor{violet!20}\textbf{ScoreNet}    & \cellcolor{violet!20}$1536 \times 2048$ &\cellcolor{violet!20} \textbf{335.0 $\pm$ 7.9}  &\cellcolor{violet!20} \xk \\ 
\bottomrule 
\end{tabular}}
\vspace{-1.5em}
\label{table:throughput}
\end{table}

\vspace{1ex} \noindent \textbf{Ablation on Shape Cues and Robustness.}
\label{par:shape_cues}
We investigate $\scorenet$’s ability to learn shape-related features. To do so, we study shape cues extracted by the recommendation model via the concatenated \texttt{[CLS]} tokens (see Fig.~\ref{fig:troi_architecture}). Consequently, we implement shape removal by applying a random permutation of the downscaled image’s tokens at test time. With this setup, a weighted F1-score of 59.8 $\pm$ 0.8\% is reached, representing a significant drop in performance compared to 64.4 $\pm$ 0.9\% without permutation. It demonstrates that \textit{i)} the recommendation stage's concatenated \texttt{[CLS]} tokens contribute positively to the overall representation and \textit{ii)} the latter is \textbf{not permutation invariant} and thus shape-dependent.  In a second experiment, we show the whole recommendation stage is also shape-dependent. To that end, we repeat the same experiment, but the patches are extracted from the permuted images, reaching a weighted F1-score of 59.5 $\pm$ 0.6\%. We further observe that for a given image, the overlap of the selected patches with and without permutation is, on average, only $15.7\%$, which indicates that the semantic distribution learned by $\scorenet$ is shape-dependent. 
\section{Conclusion and Future Work}
\label{sec:conclusion}
We have introduced $\scorenet$, an efficient transformer-based architecture that dynamically recommends discriminative regions from large histopathological images, yielding rich generalizable representations at an efficient computational cost. In addition, we propose $\scoremix$, a new attention-guided mixing augmentation that produces coherent sample-label pairs. We achieve new SOTA results on the BRACS dataset for TRoIs classification and demonstrate ScoreNet’s superior throughput improvements compared to previous SOTA efficient transformer-based architectures. 

{\small
\bibliographystyle{ieee_fullname}
\bibliography{egbib}

\begin{thebibliography}{10}\itemsep=-1pt

\bibitem{ARESTA2019122}
Guilherme Aresta, Teresa Ara{\'u}jo, Scotty Kwok, Sai~Saketh Chennamsetty,
  Mohammed Safwan, Varghese Alex, Bahram Marami, Marcel Prastawa, Monica Chan,
  Michael Donovan, et~al.
\newblock Bach: Grand challenge on breast cancer histology images.
\newblock {\em Medical image analysis}, 56:122--139, 2019.

\bibitem{aygunecs2020graph}
Bulut Ayg{\"u}ne{\c{s}}, Selim Aksoy, Ramazan~G{\"o}kberk Cinbi{\c{s}}, Kemal
  K{\"o}semehmeto{\u{g}}lu, Sevgen {\"O}nder, and Ay{\c{s}}eg{\"u}l {\"U}ner.
\newblock Graph convolutional networks for region of interest classification in
  breast histopathology.
\newblock In {\em Medical Imaging 2020: Digital Pathology}, volume 11320, page
  113200K. International Society for Optics and Photonics, 2020.

\bibitem{camelyon16}
Babak~Ehteshami Bejnordi, Mitko Veta, Paul~Johannes Van~Diest, Bram
  Van~Ginneken, Nico Karssemeijer, Geert Litjens, Jeroen~AWM Van Der~Laak,
  Meyke Hermsen, Quirine~F Manson, Maschenka Balkenhol, et~al.
\newblock Diagnostic assessment of deep learning algorithms for detection of
  lymph node metastases in women with breast cancer.
\newblock {\em Jama}, 318(22):2199--2210, 2017.

\bibitem{caron2020unsupervised}
Mathilde Caron, Ishan Misra, Julien Mairal, Priya Goyal, Piotr Bojanowski, and
  Armand Joulin.
\newblock Unsupervised learning of visual features by contrasting cluster
  assignments.
\newblock In {\em Thirty-fourth Conference on Neural Information Processing
  Systems (NeurIPS)}, 2020.

\bibitem{caron2021emerging}
Mathilde Caron, Hugo Touvron, Ishan Misra, Herv\'e J\'egou, Julien Mairal,
  Piotr Bojanowski, and Armand Joulin.
\newblock Emerging properties in self-supervised vision transformers.
\newblock In {\em Proceedings of the International Conference on Computer
  Vision (ICCV)}, 2021.

\bibitem{chen2021transmix}
Jie-Neng Chen, Shuyang Sun, Ju He, Philip Torr, Alan Yuille, and Song Bai.
\newblock Transmix: Attend to mix for vision transformers.
\newblock {\em arXiv preprint arXiv:2111.09833}, 2021.

\bibitem{chen2021multimodal}
Richard~J Chen, Ming~Y Lu, Wei-Hung Weng, Tiffany~Y Chen, Drew~FK Williamson,
  Trevor Manz, Maha Shady, and Faisal Mahmood.
\newblock Multimodal co-attention transformer for survival prediction in
  gigapixel whole slide images.
\newblock In {\em Proceedings of the IEEE/CVF International Conference on
  Computer Vision}, pages 4015--4025, 2021.

\bibitem{DBLP:journals/corr/abs-2102-10882}
Xiangxiang Chu, Bo Zhang, Zhi Tian, Xiaolin Wei, and Huaxia Xia.
\newblock Do we really need explicit position encodings for vision
  transformers?
\newblock {\em CoRR}, abs/2102.10882, 2021.

\bibitem{ciga2022self}
Ozan Ciga, Tony Xu, and Anne~Louise Martel.
\newblock Self supervised contrastive learning for digital histopathology.
\newblock {\em Machine Learning with Applications}, 7:100198, 2022.

\bibitem{cordonnier2021differentiable}
Jean-Baptiste Cordonnier, Aravindh Mahendran, Alexey Dosovitskiy, Dirk
  Weissenborn, Jakob Uszkoreit, and Thomas Unterthiner.
\newblock Differentiable patch selection for image recognition.
\newblock In {\em Proceedings of the IEEE/CVF Conference on Computer Vision and
  Pattern Recognition}, pages 2351--2360, 2021.

\bibitem{5206848}
Jia Deng, Wei Dong, Richard Socher, Li-Jia Li, Kai Li, and Li Fei-Fei.
\newblock Imagenet: A large-scale hierarchical image database.
\newblock In {\em 2009 IEEE Conference on Computer Vision and Pattern
  Recognition}, pages 248--255, 2009.

\bibitem{deng2009imagenet}
Jia Deng, Wei Dong, Richard Socher, Li-Jia Li, Kai Li, and Li Fei-Fei.
\newblock Imagenet: A large-scale hierarchical image database.
\newblock In {\em 2009 IEEE conference on computer vision and pattern
  recognition}, pages 248--255. Ieee, 2009.

\bibitem{osovitskiy2021image}
Alexey Dosovitskiy, Lucas Beyer, Alexander Kolesnikov, Dirk Weissenborn,
  Xiaohua Zhai, Thomas Unterthiner, Mostafa Dehghani, Matthias Minderer, Georg
  Heigold, Sylvain Gelly, Jakob Uszkoreit, and Neil Houlsby.
\newblock An image is worth 16x16 words: Transformers for image recognition at
  scale.
\newblock {\em ICLR}, 2021.

\bibitem{goyal2018accurate}
Priya Goyal, Piotr Dollár, Ross Girshick, Pieter Noordhuis, Lukasz Wesolowski,
  Aapo Kyrola, Andrew Tulloch, Yangqing Jia, and Kaiming He.
\newblock Accurate, large minibatch sgd: Training imagenet in 1 hour, 2018.

\bibitem{he2015delving}
Kaiming He, Xiangyu Zhang, Shaoqing Ren, and Jian Sun.
\newblock Delving deep into rectifiers: Surpassing human-level performance on
  imagenet classification.
\newblock In {\em Proceedings of the IEEE international conference on computer
  vision}, pages 1026--1034, 2015.

\bibitem{he2016deep}
Kaiming He, Xiangyu Zhang, Shaoqing Ren, and Jian Sun.
\newblock Deep residual learning for image recognition.
\newblock In {\em Proceedings of the IEEE conference on computer vision and
  pattern recognition}, pages 770--778, 2016.

\bibitem{hou2016patch}
Le Hou, Dimitris Samaras, Tahsin~M Kurc, Yi Gao, James~E Davis, and Joel~H
  Saltz.
\newblock Patch-based convolutional neural network for whole slide tissue image
  classification.
\newblock In {\em Proceedings of the IEEE conference on computer vision and
  pattern recognition}, pages 2424--2433, 2016.

\bibitem{ilse2018attention}
Maximilian Ilse, Jakub Tomczak, and Max Welling.
\newblock Attention-based deep multiple instance learning.
\newblock In {\em International conference on machine learning}, pages
  2127--2136. PMLR, 2018.

\bibitem{kalra2021pay}
Shivam Kalra, Mohammed Adnan, Sobhan Hemati, Taher Dehkharghanian, Shahryar
  Rahnamayan, and Hamid~R Tizhoosh.
\newblock Pay attention with focus: A novel learning scheme for classification
  of whole slide images.
\newblock In {\em International Conference on Medical Image Computing and
  Computer-Assisted Intervention}, pages 350--359. Springer, 2021.

\bibitem{kalra2020yottixel}
Shivam Kalra, Hamid~R Tizhoosh, Charles Choi, Sultaan Shah, Phedias Diamandis,
  Clinton~JV Campbell, and Liron Pantanowitz.
\newblock Yottixel--an image search engine for large archives of histopathology
  whole slide images.
\newblock {\em Medical Image Analysis}, 65:101757, 2020.

\bibitem{koohbanani2021self}
Navid~Alemi Koohbanani, Balagopal Unnikrishnan, Syed~Ali Khurram, Pavitra
  Krishnaswamy, and Nasir Rajpoot.
\newblock Self-path: Self-supervision for classification of pathology images
  with limited annotations.
\newblock {\em IEEE Transactions on Medical Imaging}, 2021.

\bibitem{li2021dual}
Bin Li, Yin Li, and Kevin~W Eliceiri.
\newblock Dual-stream multiple instance learning network for whole slide image
  classification with self-supervised contrastive learning.
\newblock In {\em Proceedings of the IEEE/CVF Conference on Computer Vision and
  Pattern Recognition}, pages 14318--14328, 2021.

\bibitem{li2021efficient}
Chunyuan Li, Jianwei Yang, Pengchuan Zhang, Mei Gao, Bin Xiao, Xiyang Dai, Lu
  Yuan, and Jianfeng Gao.
\newblock Efficient self-supervised vision transformers for representation
  learning.
\newblock {\em arXiv preprint arXiv:2106.09785}, 2021.

\bibitem{liu2021swin}
Ze Liu, Yutong Lin, Yue Cao, Han Hu, Yixuan Wei, Zheng Zhang, Stephen Lin, and
  Baining Guo.
\newblock Swin transformer: Hierarchical vision transformer using shifted
  windows.
\newblock In {\em Proceedings of the IEEE/CVF International Conference on
  Computer Vision}, pages 10012--10022, 2021.

\bibitem{lu2021data}
Ming~Y Lu, Drew~FK Williamson, Tiffany~Y Chen, Richard~J Chen, Matteo Barbieri,
  and Faisal Mahmood.
\newblock Data-efficient and weakly supervised computational pathology on
  whole-slide images.
\newblock {\em Nature Biomedical Engineering}, 5(6):555--570, 2021.

\bibitem{maron1998framework}
Oded Maron and Tom{\'a}s Lozano-P{\'e}rez.
\newblock A framework for multiple-instance learning.
\newblock {\em Advances in neural information processing systems}, pages
  570--576, 1998.

\bibitem{mercan2019patch}
Caner Mercan, Selim Aksoy, Ezgi Mercan, Linda~G Shapiro, Donald~L Weaver, and
  Joann~G Elmore.
\newblock From patch-level to roi-level deep feature representations for breast
  histopathology classification.
\newblock In {\em Medical Imaging 2019: Digital Pathology}, volume 10956, page
  109560H. International Society for Optics and Photonics, 2019.

\bibitem{paszke2019pytorch}
Adam Paszke, Sam Gross, Francisco Massa, Adam Lerer, James Bradbury, Gregory
  Chanan, Trevor Killeen, Zeming Lin, Natalia Gimelshein, Luca Antiga, et~al.
\newblock Pytorch: An imperative style, high-performance deep learning library.
\newblock {\em Advances in neural information processing systems},
  32:8026--8037, 2019.

\bibitem{pati2020hact}
Pushpak Pati, Maria Frucci, and Maria Gabrani.
\newblock Hact-net: A hierarchical cell-to-tissue graph neural network for
  histopathological image classification.
\newblock In {\em Uncertainty for Safe Utilization of Machine Learning in
  Medical Imaging, and Graphs in Biomedical Image Analysis: Second
  International Workshop, UNSURE 2020, and Third International Workshop, GRAIL
  2020, Held in Conjunction with MICCAI 2020, Lima, Peru, October 8, 2020,
  Proceedings}, volume 12443, page 208. Springer Nature, 2020.

\bibitem{pati2021hierarchical}
Pushpak Pati, Guillaume Jaume, Antonio Foncubierta, Florinda Feroce, Anna~Maria
  Anniciello, Giosu{\`e} Scognamiglio, Nadia Brancati, Maryse Fiche, Estelle
  Dubruc, Daniel Riccio, et~al.
\newblock Hierarchical cell-to-tissue graph representations for breast cancer
  subtyping in digital pathology.
\newblock {\em arXiv e-prints}, pages arXiv--2102, 2021.

\bibitem{PATI2022102264}
Pushpak Pati, Guillaume Jaume, Antonio Foncubierta-Rodríguez, Florinda Feroce,
  Anna~Maria Anniciello, Giosue Scognamiglio, Nadia Brancati, Maryse Fiche,
  Estelle Dubruc, Daniel Riccio, Maurizio {Di Bonito}, Giuseppe {De Pietro},
  Gerardo Botti, Jean-Philippe Thiran, Maria Frucci, Orcun Goksel, and Maria
  Gabrani.
\newblock Hierarchical graph representations in digital pathology.
\newblock {\em Medical Image Analysis}, 75:102264, 2022.

\bibitem{romero2020group}
David~W Romero and Jean-Baptiste Cordonnier.
\newblock Group equivariant stand-alone self-attention for vision.
\newblock In {\em International Conference on Learning Representations}, 2020.

\bibitem{rymarczyk2021kernel}
Dawid Rymarczyk, Adriana Borowa, Jacek Tabor, and Bartosz Zielinski.
\newblock Kernel self-attention for weakly-supervised image classification
  using deep multiple instance learning.
\newblock In {\em Proceedings of the IEEE/CVF Winter Conference on Applications
  of Computer Vision}, pages 1721--1730, 2021.

\bibitem{shao2021transmil}
Zhuchen Shao, Hao Bian, Yang Chen, Yifeng Wang, Jian Zhang, Xiangyang Ji, and
  Yongbing Zhang.
\newblock Transmil: Transformer based correlated multiple instance learning for
  whole slide image classication.
\newblock {\em arXiv preprint arXiv:2106.00908}, 2021.

\bibitem{singh2020don}
Krishna~Kumar Singh, Dhruv Mahajan, Kristen Grauman, Yong~Jae Lee, Matt
  Feiszli, and Deepti Ghadiyaram.
\newblock Don't judge an object by its context: Learning to overcome contextual
  bias.
\newblock In {\em Proceedings of the IEEE/CVF Conference on Computer Vision and
  Pattern Recognition}, pages 11070--11078, 2020.

\bibitem{sirinukunwattana2018improving}
Korsuk Sirinukunwattana, Nasullah~Khalid Alham, Clare Verrill, and Jens
  Rittscher.
\newblock Improving whole slide segmentation through visual context-a
  systematic study.
\newblock In {\em International Conference on Medical Image Computing and
  Computer-Assisted Intervention}, pages 192--200. Springer, 2018.

\bibitem{Srinidhi2021deep}
Chetan~L Srinidhi, Ozan Ciga, and Anne~L Martel.
\newblock Deep neural network models for computational histopathology: A
  survey.
\newblock {\em Medical Image Analysis}, 67:101813, 2021.

\bibitem{srinidhi2021self}
Chetan~L Srinidhi, Seung~Wook Kim, Fu-Der Chen, and Anne~L Martel.
\newblock Self-supervised driven consistency training for annotation efficient
  histopathology image analysis.
\newblock {\em arXiv preprint arXiv:2102.03897}, 2021.

\bibitem{tokunaga2019adaptive}
Hiroki Tokunaga, Yuki Teramoto, Akihiko Yoshizawa, and Ryoma Bise.
\newblock Adaptive weighting multi-field-of-view cnn for semantic segmentation
  in pathology.
\newblock In {\em Proceedings of the IEEE/CVF Conference on Computer Vision and
  Pattern Recognition}, pages 12597--12606, 2019.

\bibitem{uddin2020saliencymix}
AFM~Shahab Uddin, Mst~Sirazam Monira, Wheemyung Shin, TaeChoong Chung, and
  Sung-Ho Bae.
\newblock Saliencymix: A saliency guided data augmentation strategy for better
  regularization.
\newblock In {\em International Conference on Learning Representations}, 2020.

\bibitem{vaswani2017attention}
Ashish Vaswani, Noam Shazeer, Niki Parmar, Jakob Uszkoreit, Llion Jones,
  Aidan~N Gomez, {\L}ukasz Kaiser, and Illia Polosukhin.
\newblock Attention is all you need.
\newblock In {\em Advances in neural information processing systems}, pages
  5998--6008, 2017.

\bibitem{walawalkar2020attentive}
Devesh Walawalkar, Zhiqiang Shen, Zechun Liu, and Marios Savvides.
\newblock Attentive cutmix: An enhanced data augmentation approach for deep
  learning based image classification.
\newblock In {\em ICASSP 2020-2020 IEEE International Conference on Acoustics,
  Speech and Signal Processing (ICASSP)}, pages 3642--3646. IEEE, 2020.

\bibitem{wang2021transpath}
Xiyue Wang, Sen Yang, Jun Zhang, Minghui Wang, Jing Zhang, Junzhou Huang, Wei
  Yang, and Xiao Han.
\newblock Transpath: Transformer-based self-supervised learning for
  histopathological image classification.
\newblock In {\em International Conference on Medical Image Computing and
  Computer-Assisted Intervention}, pages 186--195. Springer, 2021.

\bibitem{xu2015deep}
Yan Xu, Zhipeng Jia, Yuqing Ai, Fang Zhang, Maode Lai, I Eric, and Chao Chang.
\newblock Deep convolutional activation features for large scale brain tumor
  histopathology image classification and segmentation.
\newblock In {\em 2015 IEEE international conference on acoustics, speech and
  signal processing (ICASSP)}, pages 947--951. IEEE, 2015.

\bibitem{yun2019cutmix}
Sangdoo Yun, Dongyoon Han, Seong~Joon Oh, Sanghyuk Chun, Junsuk Choe, and
  Youngjoon Yoo.
\newblock Cutmix: Regularization strategy to train strong classifiers with
  localizable features.
\newblock In {\em Proceedings of the IEEE/CVF International Conference on
  Computer Vision}, pages 6023--6032, 2019.

\bibitem{zhang2018mixup}
Hongyi Zhang, Moustapha Cisse, Yann~N. Dauphin, and David Lopez-Paz.
\newblock mixup: Beyond empirical risk minimization.
\newblock {\em International Conference on Learning Representations}, 2018.

\bibitem{zhou2019cgc}
Yanning Zhou, Simon Graham, Navid Alemi~Koohbanani, Muhammad Shaban, Pheng-Ann
  Heng, and Nasir Rajpoot.
\newblock Cgc-net: Cell graph convolutional network for grading of colorectal
  cancer histology images.
\newblock In {\em Proceedings of the IEEE/CVF International Conference on
  Computer Vision Workshops}, pages 0--0, 2019.

\end{thebibliography}
}
\clearpage
\appendix
\section{Appendix Overview}
\label{sec:overview}
In this appendix, we provide additional ablation studies and experimental details. The remaining of this appendix is organized as follows. In Sec.~\ref{sec:exp_setup} we detail the architectural and training details, e.g., parameters choices. Additional ablations are detailed in Sec.~\ref{sec:add_ablations} . A detailed derivation of the computational cost is presented in Sec.~\ref{sec:comp_cost}. We discuss, in Sec.~\ref{sec:scoremix_investigation}, some properties of $\scoremix$ and present some examples of our proposed $\scoremix$ augmentation. Finally, the suitability of $\scorenet$ to learn from uncurated data is evaluated in Sec.~\ref{sec:uncurated_learning}.

\section{Experimental Setup \& Datasets}
\label{sec:exp_setup}

\subsection{Networks Architectures}
\label{subsec:nets_architectures}
\vspace{1ex}\noindent \textbf{ScoreNet.}
The proposed $\scorenet$ architecture comprises two stages: the recommendation and aggregation stages. The former leverages a modified $\vit$-Tiny to produces the semantic distribution. Similarly, the latter relies on an identical $\vit$-Tiny to independently embed the selected high-resolution patches (\textit{local fine-grained attention}) and on a transformer encoder to mix the embedded patches (\textit{global coarse-grained attention}). The following parameters of the two identical $\vit$-Tiny were modified to be tailored for the task:

 \begin{itemize}
     \setlength\itemsep{-0.2em}
     \item \texttt{embed\_dim=96}.
     \item \texttt{depth=8}.
     \item \texttt{num\_heads=4}.
     \item  \texttt{mlp\_ratio=2}.
 \end{itemize}
 These modifications were brought to allow for a self-supervised pre-training with a sufficiently large batch size ($bs\geq 128$), which was reported to be of significant importance to reach good performance \cite{caron2021emerging}. The parameters of the transformer encoder implementing the \textit{global coarse-grained attention} mechanism are:
 \begin{itemize}
     \setlength\itemsep{-0.2em}
     \item \texttt{embed\_dim=96}.
     \item \texttt{depth=4}.
     \item \texttt{num\_heads=4}.
     \item  \texttt{mlp\_ratio=2}.
 \end{itemize}
 Overall $\scorenet$'s model totals approximately $1.79$M parameters. 
 
\vspace{1ex}\noindent \textbf{SwinTransformer.}
SwinTransformers \cite{liu2021swin} relies on hierarchical architecture attention mechanism, namely intra- and inter-window attentions. The patch-merging operation reduces the time, and memory cost of SwinTransformers \cite{liu2021swin} significantly, which decreases the total number of tokens by $4$, while increasing the embedding by $2$. The architecture is modified to accept non-square windows, allowing SwinTransformers to process non-square images
images. The resulting parameters are:

\begin{itemize}
     \setlength\itemsep{-0.2em}
    \item \texttt{patch\_size=16}.
    \item \texttt{input\_embed\_dim\_size=24}.
    \item \texttt{output\_embed\_dim\_size=192}.
    \item \texttt{depths=[2, 2, 6, 2]}.
    \item \texttt{num\_heads=[3, 6, 12, 24]}.
    \item  \texttt{window\_size=(6, 8)}.
    \item  \texttt{mlp\_ratio=4}.
\end{itemize}
 Overall the SwinTransformer model totals approximately $1.77$M parameters. 

\vspace{1ex}\noindent \textbf{TransPath.}
As described in \cite{wang2021transpath}, TransPath's architecture leverages a CNN encoder to jointly reduce the input image's size, extract relevant features, and tile the image in pre-embedded patches. Subsequently, a transformer encoder processes the CNN encoder's features to capture global interactions. The CNN encoder's architecture is as follows:
 \begin{itemize}
     \setlength\itemsep{-0.2em}
     \item \texttt{n\_convolutions=4}.
     \item \texttt{n\_filters=[8, 32, 128, 512]}.
     \item \texttt{kernel\_sizes=[(3, 3), (3, 3), (3, 3), (3, 3)]}.
     \item \texttt{pooling\_kernel\_sizes=[(4, 4), (2, 2), (4, 4), (4, 4)]}.
     \item \texttt{activation=ReLU} \cite{he2015delving}.
 \end{itemize}
A projection convolution is used to match the desired embedding dimension of the transformer encoder. Its parameters are:

\begin{itemize}
    \item \texttt{n\_filters=192}.
    \item \texttt{kernel\_sizes=(1, 1)}.
\end{itemize}
The parameters of the transformer encoder are:
 \begin{itemize}
     \setlength\itemsep{-0.2em}
     \item \texttt{embed\_dim=192}.
     \item \texttt{depth=4}.
     \item \texttt{num\_heads=4}.
     \item  \texttt{mlp\_ratio=2}.
 \end{itemize}
 Each transformer block rely on TransPath's customized token-aggregating and excitation multi-head self-attention (MHSA-TAE) \cite{wang2021transpath}.
 Overall, TransPath's model totals approximately $1.93$M parameters. 
 
\vspace{1ex}\noindent \textbf{TransMIL.}
We adopt the original implementation as provided by the authors \cite{shao2021transmil}. It relies on a ResNet-50 \cite{he2016deep} pre-trained on ImageNet \cite{5206848} to embed the individual $256\times256$ patches. 
Overall, TransMIL's model totals approximately $3.19$M parameters (not counting the parameters of the ResNet-50). %

\vspace{1ex}\noindent \textbf{CLAM.}
The implementation of CLAM follows the code provided by the authors \cite{lu2021data}. It relies on a ResNet-50 \cite{he2016deep} pre-trained on ImageNet \cite{5206848} to embed the individual $256\times256$ patches. 
Overall, the variations of CLAM-(SB/MB)/(S/B) total from $1.32$M to $1.46$M parameters (not counting the parameters of the ResNet-50). 
 
\subsection{Self-Supervised Pre-training}
\label{subsec:ss_pretraining}
\noindent \textbf{Modular Pre-training.}
Our modular architecture allows for independent self-supervised pre-trainings of the recommendation stage's $\vit$ and that of the local fine-grained attention mechanism. A two steps pre-training can be beneficial, as it provides the possibility to validate each part independently. Similarly, one of the modules, typically the one implementing the fine-grained local attention, can be trained on an auxiliary annotated dataset or be replaced by a standard pre-trained model. 

The self-supervised pre-training follows the guidelines of \cite{caron2021emerging}. Apart from the differences in architectures described in Sec.~\ref{subsec:nets_architectures}, minor modifications were made in the projection head to account for the reduced heterogeneity in our datasets compared to that in ImageNet \cite{deng2009imagenet}. The modifications are:
\begin{itemize}
    \setlength\itemsep{-0.2em}
    \item \texttt{hidden\_dim=1024}.
    \item \texttt{bottleneck\_dim=128}.
    \item \texttt{out\_dim=1024}.
\end{itemize}
These modifications are in line with the interpretation of \cite{caron2020unsupervised} which considers the last linear layer as a projection on a set of learnable centroids and that their number should reflect the level of diversity present in the dataset. For this interpretation to hold, it is required that both the last layer's input and its weights are normalized, which is the case in our setup. The remaining parameters, aside from the position encoding which is discussed in Sec.~\ref{par:pos_enc}, are set to the default values (see \cite{caron2021emerging} for details).

\vspace{1ex}\noindent \textbf{End-to-end Pre-training.}
In some cases, an end-to-end pre-training of $\scorenet$ is preferable. For that purpose, we experimented with two approaches: DINO and a variant of it for that purpose. The former uses the default values for all parameters but those of the projection head described above. On the contrary, the latter benefits from different augmentations and another pretext task and thereby avoid a potential pitfall of DINO: encouraging contextual bias \cite{singh2020don}, which occurs when the similarity between the representations of views depicting distinct tissue types is enforced.

In this regard, the set of admissible augmentations are constrained to those that change the pixels' values, but not their locations. Consequently, a given image's different views are bounded to bear identical semantic content. A key aspect of DINO's strong performance is due to enforcing the local-to-global correspondence between the student's local crops and that of the teacher's global crop. To mimic that knowledge distillation mechanism, we encourage the student, which only processes the most discriminative high-resolution patches, to match the teacher's representation, which on the contrary, is based on all high-resolution patches. One can observe that this pretext task enforces local-to-global correspondence while providing a strong supervisory signal to the student's scorer, which has to highlight the most relevant regions for the task to be successfully accomplished. 

In that setting, $\scorenet$'s representation is obtained by the concatenation of the \texttt{[CLS]} tokens of the \textit{global coarse-grained attention} module's last two transformer blocks. This representation benefits from global contextual information through the teacher, which processes the whole high-resolution image. The projection head's parameters are identified as described above.

\subsection{Datasets}
\label{subsec:dataset}
In addition to the annotated $\trois$ from two datasets, namely BRACS \cite{pati2021hierarchical} and BACH \cite{ARESTA2019122}, additional sets of unlabeled of images are used to pre-train the models and for various ablations. The sets of unlabeled images are detailed here.
\par
\vspace{1ex}\noindent \textbf{BRACS.}
The BRACS dataset encompasses both the annotated $\trois$ and the 547 whole-slide images from which they were extracted. We use the $\wsis$ to create an unlabeled pre-training set. More precisely, two types of auxiliary datasets are extracted from BRACS's $\wsis$: tiles set at $40\times$ and low-resolution thumbnails set at $\frac{40}{s}\times$, where $s$ is the down-scaling ratio. The former set is used to pre-train the \textit{local fine-grained attention} module, whereas the latter serves to pre-train the recommendation stag's scorer. We experimented with two variants of these paired sets. The first variant is designed for a version of $\scorenet$, where the dimension of the finely attended regions is $P_{h}=224$, the recommendation stage processes low-resolution patches of dimension $P_{l}=16$ and consequently a down-scaling ratio $s=14$. The second variant is designed for a version of $\scorenet$, where the dimension of the finely attended regions is $P_{h}=128$, the recommendation stage processes low-resolution patches of dimension $P_{l}=16$ and consequently a down-scaling ratio $s=8$. The resulting sets contain approximately $150$k images (for a fair comparison of the two versions, see Sec.~\ref{par:patch_dimensions}). 

The last images are extracted from BRACS to conduct TransPath's self-supervised pre-training. From the $\wsis$, an unlabeled set of approximately $100$k images at $40\times$ are extracted. The images have dimensions $1536\times1536$, which is approximately the median dimensions of the annotated $\trois$. 
\par
\vspace{1ex}\noindent \textbf{BACH.}
Similarly, the BACH dataset comprises an annotated set of $\trois$ and the accompanying $40$ whole-slide images. From the $\wsis$, an unlabeled pre-training set of approximately $11$k images at $20\times$ are extracted. The images have the exact dimensions as the annotated $\trois$, $1536\times2048$.
 \par
\vspace{1ex}\noindent \textbf{CAMELYON16.}
Finally, additional tiles set is extracted from CAMELYON16, which is, to our knowledge, the only one with patch-level annotations. This set is used to evaluate the pre-training of the fine-grained attention module. The latter is composed of 10k images at $40\times$, of dimensions $128\times128$ or $224\times224$. It is class-balanced, and any patch which contains tumorous tissue is considered tumour positive. This set is also used to measure the effectiveness of the position encoding on the fine-grained attention module in Sec.~\ref{par:pos_enc}. \par

\section{Additional Ablations}
\label{sec:add_ablations}

\noindent\textbf{Down-Scaling Ratio \& Dimensions of the Attended Regions.}
A key component of the proposed pipeline is to determine the down-scaling ratio, $s$, and the dimension of the square patches in low-resolution, $P_{l} \times P_{l}$, and in high-resolution, $P_{h} \times P_{h}$. Considering the well-studied nature of the $\vit$s scorers, we use the standard patch dimension $P_{l} = 16$ for the patches in low-resolution. It has been shown that smaller patches ($P_{l} = 8$ or $P_{l} = 5$) improve the quality of the learned representations \cite{caron2021emerging}, nonetheless the incurred increase in computational and memory cost is unsuitable for our application. For the high-resolution patches, we experiment with two standard patch dimensions: $P_{h} = 128$ and $P_{h} = 224$. As the self-attention of the recommendation stage is used as a learnable distribution of the semantic content, there should exist a 1-to-1 mapping between the low-resolution patches and the high-resolution regions that can be extracted. As a consequence, the down-scaling ratio is fully determined by the dimensions of the patches: $s = P_{h} / P_{l}$. In our case, it translates to down-scaling ratio of either $s = 8$, or $s = 14$. 
\label{par:patch_dimensions}
\begin{table*}[ht]
\centering
\caption{\textbf{A weighted $k$ Nearest Neighbors classifier assesses the learned features' discriminability (weighted F1-score) on the low-resolution BACH dataset}. The performances of CNN and $\vit$-based architectures are compared, and similarly for two down-scaling ratios ($s=8$ or $s=14$). We use a 4-fold scheme with $75\%/25\%$ train/test splits.}
\resizebox{1\textwidth}{!}{
\begin{tabular}{l c c c c c c c c}
\hline
       & \multicolumn{4}{c}{ViT} & \multicolumn{4}{c}{CNN}  \\
       \cmidrule(lr){2-5} \cmidrule(lr){6-9}
       & \multicolumn{2}{c}{Teacher} & \multicolumn{2}{c}{Student} & \multicolumn{2}{c}{Teacher} & \multicolumn{2}{c}{Student}\\
\cmidrule(lr){2-3} \cmidrule(lr){4-5} \cmidrule(lr){6-7} \cmidrule(lr){8-9}
$k$ & $s=14$ & $s=8$  & $s=14$ & $s=8$  & $s=14$ & $s=8$ & $s=14$ & $s=8$ \\
\hline
 1   & 71.7 $\pm$ 6.4 & \textbf{78.5 $\pm$ 6.4} & 73.6 $\pm$ 5.1 & 77.4 $\pm$ 5.1 & 63.6 $\pm$ 5.1 & 64.4 $\pm$ 1.9 & 63.8 $\pm$ 3.2 & 63.9 $\pm$ 2.2 \\
 5   & 71.5 $\pm$ 1.7 & \textbf{81.7 $\pm$ 3.2} & 72.8 $\pm$ 1.9 & 81.0 $\pm$ 4.0 & 65.1 $\pm$ 3.3 & 64.7 $\pm$ 2.1 & 64.1 $\pm$ 4.6 & 65.4 $\pm$ 2.7 \\
 10  & 71.9 $\pm$ 2.4 & 77.8 $\pm$ 2.8 & 72.5 $\pm$ 2.5 & \textbf{77.9 $\pm$ 3.4} & 62.0 $\pm$ 3.8 & 58.9 $\pm$ 2.9 & 61.5 $\pm$ 5.8 & 61.1 $\pm$ 2.5 \\
 20  & 71.3 $\pm$ 4.0 & 76.3 $\pm$ 3.0 & 72.5 $\pm$ 3.0 & \textbf{76.5 $\pm$ 4.0} & 64.0 $\pm$ 6.7 & 55.5 $\pm$ 1.8 & 61.0 $\pm$ 9.2 & 55.4 $\pm$ 2.4 \\
 50  & 71.2 $\pm$ 4.0 & \textbf{74.7 $\pm$ 4.7} & 70.9 $\pm$ 3.3 & 74.3 $\pm$ 5.7 & 59.3 $\pm$ 5.3 & 56.1 $\pm$ 3.2 & 58.1 $\pm$ 6.2 & 54.6 $\pm$ 3.9 \\
 100 & 71.7 $\pm$ 4.1 & \textbf{74.0 $\pm$ 5.5} & 71.4 $\pm$ 3.8 & 73.6 $\pm$ 5.9 & 57.4 $\pm$ 3.4 & 50.6 $\pm$ 5.1 & 56.2 $\pm$ 3.0 & 48.7 $\pm$ 4.7 \\
\hline
\end{tabular}
}
\vspace{0.2cm}
\label{table:knn_vits_cnns_bach}
\end{table*}

\begin{table}[ht]
\centering
\caption{\textbf{A weighted $k$ Nearest Neighbors classifier assesses the learned features' discriminability (weighted F1-score) on the low-resolution BRACS dataset}. The performances of CNN and $\vit$-based architectures are compared, and similarly for two down-scaling ratios ($s=8$ or $s=14$). The $k$-NN classifier is trained on the merged train/valid set and evaluated on the test set (see \cite{pati2021hierarchical}), hence the high performances.}
\resizebox{\columnwidth}{!}{
\begin{tabular}{l c c c c c c c c}
\hline
       & \multicolumn{4}{c}{ViT} & \multicolumn{4}{c}{CNN}  \\
       \cmidrule(lr){2-5} \cmidrule(lr){6-9}
       & \multicolumn{2}{c}{Teacher} & \multicolumn{2}{c}{Student} & \multicolumn{2}{c}{Teacher} & \multicolumn{2}{c}{Student}\\
\cmidrule(lr){2-3} \cmidrule(lr){4-5} \cmidrule(lr){6-7} \cmidrule(lr){8-9}
$k$ & $s=14$ & $s=8$  & $s=14$ & $s=8$  & $s=14$ & $s=8$ & $s=14$ & $s=8$ \\
\hline
1    & 52.5 & 54.3 & 51.6 & \textbf{55.0} & 45.2 & 45.5 & 45.4 & 44.7\\
5    & 55.2 & \textbf{56.1} & 55.4 & 55.8 & 47.1 & 47.6 & 46.6 & 46.2\\
10   & 57.2 & 56.4 & \textbf{57.5} & 56.7 & 49.3 & 46.5 & 50.5 & 45.8\\
20   & 56.9 & 58.0 & \textbf{58.1} & 57.6 & 47.1 & 47.6 & 45.9 & 47.0\\
50   & 56.2 & \textbf{57.5} & 55.7 & 56.9 & 41.2 & 44.9 & 40.6 & 44.9\\
100  & 53.9 & 54.0 & \textbf{54.3} & 53.7 & 40.3 & 43.5 & 40.1 & 44.2\\
\hline
\end{tabular}}
\vspace{0.2cm}
\label{table:knn_vits_cnns_bracs}
\end{table}


\begin{table*}[ht]
\centering
\caption{\textbf{A weighted $k$ Nearest Neighbors classifier assesses the discriminability (weighted F1-score) of the learned features on the tile CAMELYON16 dataset} (see Sec.~\ref{subsec:dataset}). The performances of CNN and $\vit$-based architectures is compared, and similarly for two tile dimensions ($128\times 128$ and $224\times 224$) corresponding to down-scaling ratios of $s=8$ and $s=14$, respectively. A 4-fold approach with $75\%/25\%$ train/test splits is used.}
\resizebox{1\textwidth}{!}{
\begin{tabular}{l c c c c c c c c}
\hline
       & \multicolumn{4}{c}{ViT} & \multicolumn{4}{c}{CNN}  \\
       \cmidrule(lr){2-5} \cmidrule(lr){6-9}
       & \multicolumn{2}{c}{Teacher} & \multicolumn{2}{c}{Student} & \multicolumn{2}{c}{Teacher} & \multicolumn{2}{c}{Student}\\
\cmidrule(lr){2-3} \cmidrule(lr){4-5} \cmidrule(lr){6-7} \cmidrule(lr){8-9}
$k$ & $s=14$ & $s=8$  & $s=14$ & $s=8$  & $s=14$ & $s=8$ & $s=14$ & $s=8$ \\
\hline
1    & 89.7 $\pm$ 0.6 & 89.1 $\pm$ 0.4 & \textbf{89.6 $\pm$ 0.6} & 89.1 $\pm$ 0.4 & 87.0 $\pm$ 0.8 & 85.8 $\pm$ 1.1 & 87.2 $\pm$ 0.9 & 85.8 $\pm$ 0.8\\
5    & \textbf{91.7 $\pm$ 0.4} & 91.1 $\pm$ 0.5 & 91.6 $\pm$ 0.3 & 91.2 $\pm$ 0.6 & 89.9 $\pm$ 1.7 & 88.8 $\pm$ 1.8 & 89.8 $\pm$ 1.7 & 88.7 $\pm$ 1.7\\
10   & \textbf{91.9 $\pm$ 0.5} & 91.4 $\pm$ 0.5 & \textbf{91.9 $\pm$ 0.5} & 91.5 $\pm$ 0.4 & 90.3 $\pm$ 1.0 & 89.0 $\pm$ 0.5 & 90.2 $\pm$ 1.1 & 89.0 $\pm$ 0.6\\
20   & \textbf{91.6 $\pm$ 0.6} & 91.2 $\pm$ 0.3 & \textbf{91.6 $\pm$ 0.4} & 91.2 $\pm$ 0.4 & 90.0 $\pm$ 1.1 & 89.0 $\pm$ 0.5 & 89.8 $\pm$ 1.1 & 88.9 $\pm$ 0.6\\
50   & \textbf{91.4 $\pm$ 0.9} & 90.7 $\pm$ 0.6 & 91.3 $\pm$ 1.0 & 90.7 $\pm$ 0.5 & 88.8 $\pm$ 1.1 & 88.6 $\pm$ 0.8 & 88.9 $\pm$ 1.1 & 88.5 $\pm$ 0.8\\
100  & \textbf{90.9 $\pm$ 1.1} & 90.1 $\pm$ 0.6 & \textbf{90.9 $\pm$ 1.1} & 90.0 $\pm$ 0.5 & 88.2 $\pm$ 1.0 & 87.6 $\pm$ 1.0 & 88.1 $\pm$ 1.0 & 87.6 $\pm$ 0.9\\
\hline
\end{tabular}
}
\vspace{0.2cm}
\label{table:knn_vits_cnn_camelyon}
\end{table*}


To find out which of these two setups is the most suitable for our application, we compare the models obtained by each of them via a weighted $k$ Nearest Neighbours classifier, which has the advantage of being fast and not requiring any finetuning. In Table~\ref{table:knn_vits_cnns_bracs}, we compare the classification results on the low-resolution ($\frac{40}{s}\times$) BRACS dataset. We report the results of both the teacher and the student models as well as those obtained by a CNN with comparable capacity and identical pre-training. We do not observe significant differences between the two scales. On the other hand, these differences are much more emphasized when evaluating the same models on the low-resolution ($\frac{20}{s}\times$) BACH dataset (see Table~\ref{table:knn_vits_cnns_bach}). These promising results on the BACH dataset, despite the mismatched scales, are to be credited to the local to global views pre-training method \cite{caron2021emerging}.

\begin{figure*}
    \centering
    \includegraphics[width=0.8\textwidth]{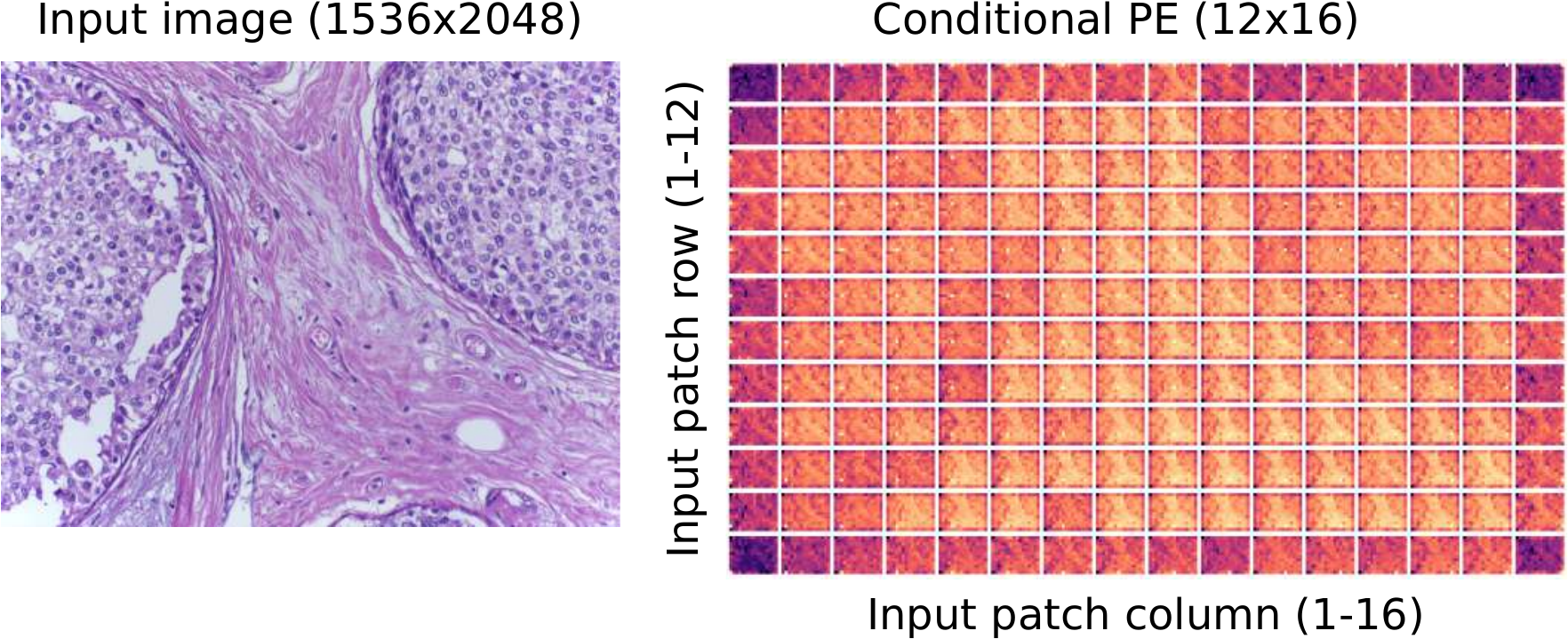}
    \caption{\textbf{The conditional position encoding \cite{DBLP:journals/corr/abs-2102-10882} of a non-squared input image} is represented. The PE is image-dependent and captures the local interactions between tokens.}
    \label{fig:conditional_pe}
\end{figure*}

The quality of the fine-grained attention module is assessed with the aforementioned method on the tile CAMELYON16 dataset introduced in Sec.~\ref{subsec:dataset}, and the hereby obtained results are reported in Table \ref{table:knn_vits_cnn_camelyon}. In conclusion, we observe that the difference is either marginal (Table \ref{table:knn_vits_cnns_bracs} \& \ref{table:knn_vits_cnn_camelyon}) or significantly in favor of the setup where $s=8$ (Table \ref{table:knn_vits_cnns_bach}) and therefore we choose this setup for the remaining experiments and architectures. As a side note, the CNN architecture performs substantially worth, but it is most likely due to the fact that the DINO \cite{caron2021emerging} method is biased towards $\vit$ architectures.
\begin{figure*}
    \centering
    \includegraphics[width=\textwidth]{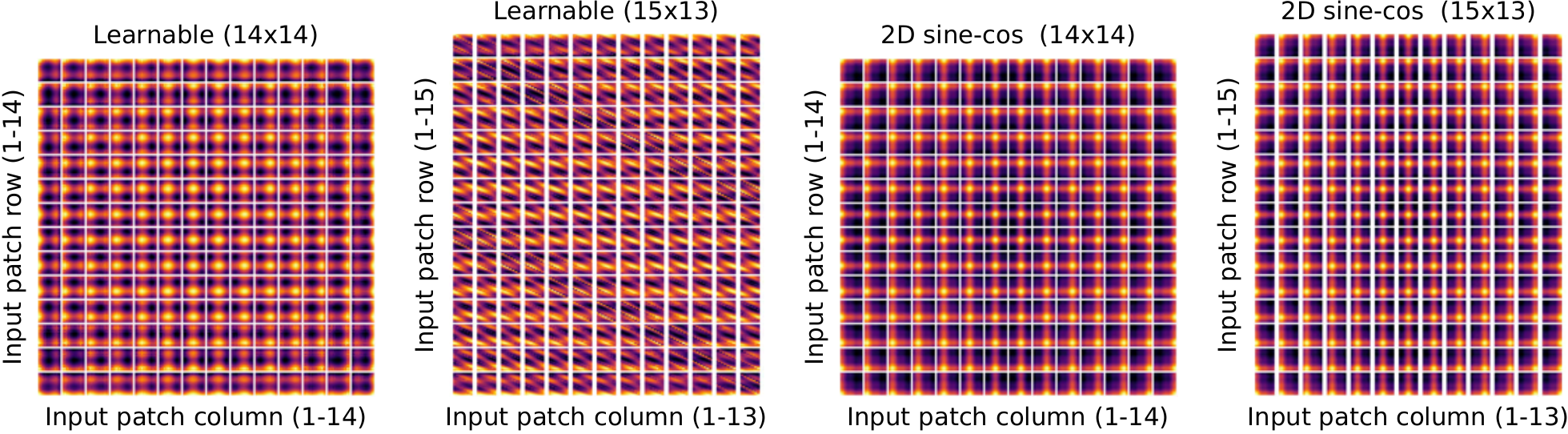}
    \caption{\textbf{The cosine similarity of a learnable and 2D sine-cos positional encoding is compared}. The learnable positional encoding introduces undesirable artifacts when the aspect ratio changes (\textit{Learnable (15$\times$13)}).}
    \label{fig:absolute_pe}
\end{figure*}
\noindent\textbf{Positional Encoding.}
\label{par:pos_enc} Without position encoding (PE), a $\vit$ processes tokens as a set and hence completely discards the global shape information; therefore, position encoding is essential. The typical approach is to learn a single matrix of absolute and additive position encoding jointly during the training phase. This approach suffers from two drawbacks: \textit{i)} the absolute encoding of each token's position implies that a pattern is different at every location it occurs, which reduces the sample efficiency \cite{romero2020group}, and \textit{ii)} as a consequence of the storage of the position encoding in a single matrix, the model treats the input tokens as a 1D sequence and thus mislays the multi-dimensionality of the inputs. The latter is not an issue as long as the input images have the same aspect ratio, as is the case with the local/global crops strategy of DINO \cite{caron2021emerging}. Nonetheless, and as depicted in Fig.~\ref{fig:absolute_pe}, this approach fails when the model is fed an image of a different aspect ratio than those used to train the position encoding.
As illustrated in Fig.~\ref{fig:absolute_pe}, the 2D sine-cos position encoding does not introduce any artifacts when used with images of different resolutions. On the other hand, any absolute position encoding is not a translation equivariant operation, an undesired property for planar images. For these reasons, we experiment with Conditional Position encoding Vision Transformer (CPVT) \cite{DBLP:journals/corr/abs-2102-10882}. This PE is input-dependent and convolution-based; consequently, it is suitable for any input resolution and patch-wise translation-equivariant. Fig.~\ref{fig:conditional_pe} reveals that the PE of border tokens is slightly different due to the needed zero-padding. This finding suggests that the absolute position encoding can be inferred from zero-paddings \cite{DBLP:journals/corr/abs-2102-10882}. We argue that CPVT is well suited to be used conjointly with $\scoremix$ as the local processing of the token is convenient for detecting local discontinuity caused by the pasting operation, which is needed to incorporate the added content to the global representation (see Sec.~\ref{sec:scoremix_investigation}).
In Table~\ref{table:pe_knn_bracs_bach_s8} and Table~\ref{table:pe_knn_camelyon_128}, we evaluate the discriminability of the features obtained by a pre-training under the DINO framework and with various position encoding methods. Table~\ref{table:pe_knn_camelyon_128}, which reports results on the tile CAMELYON16 dataset (see Sec.~\ref{subsec:dataset}), does not provide substantial shreds of evidence in favor of one PE or the other; we postulate that this lack of significant differences is due to the lessened importance of position encoding for the tile dataset. 
Indeed, at $40\times$ and with tiles of dimension $128\times 128$, the available features are mostly texture-based, and the relative organization of the patches is less relevant. This claim is well supported by the substantial differences in performance obtained by distinct PE when evaluated on the low-resolution BACH and BRACS datasets (see Table~\ref{table:pe_knn_camelyon_128}). 
These differences are further exacerbated by the fact that images on which performance is evaluated are either of varied size (BRACS) or at least of a different dimension than those used during the pre-training (BACH). Notably, there seems to be a significant performance discrepancy between the models using a [CLS] token (CPVT) and those based on a global average pooling (CPVT-GAP). Based on these results, we select the CPVT-GAP approach for the remaining experiments. Note that we referred to [CLS] token throughout this text when referring to a GAP token. Additionally, we have slightly modified the method to be able to extract one self-attention map per transformer head: instead of performing the GAP operation after the very last layer of the transformer encoder, we do it after the $(L-1)^{th}$ layer and concatenate the resulting token to the sequence, thereby producing a pseudo [CLS] token. Similarly, when $m$ pseudo [CLS] tokens are used, this operation is performed after the $(L-m)^{th}$ layer.
\begin{table*}[ht]
\centering
\caption{\textbf{A weighted $k$ Nearest Neighbors classifier assesses the learned features' discriminability (weighted F1-score) on the low-resolution BACH and BRACS datasets}.  A fixed and absolute PE (2D sine-cos)'s performances are compared to a learnable and conditional PE (CPVT and CPVT-GAP). The $k$-NN classifier is trained on the merged train/valid set and evaluated on the test set (BRACS), and a 4-fold approach with $75\%/25\%$ train/test splits is used for BACH dataset.}
\resizebox{\textwidth}{!}{
\begin{tabular}{l c c c c c c | c c c c c c}
\hline
       & \multicolumn{6}{c}{BACH} & \multicolumn{6}{c}{BRACS}  \\
       \cmidrule(lr){2-7} \cmidrule(lr){8-13}
       & \multicolumn{2}{c}{2D sine-cos} & \multicolumn{2}{c}{CPVT} & \multicolumn{2}{c}{CPVT-GAP} & \multicolumn{2}{c}{2D sine-cos} & \multicolumn{2}{c}{CPVT} & \multicolumn{2}{c}{CPVT-GAP}\\
\cmidrule(lr){2-3} \cmidrule(lr){4-5} \cmidrule(lr){6-7} \cmidrule(lr){8-9} \cmidrule(lr){10-11} \cmidrule(lr){12-13}
$k$  & Teacher & Student & Teacher & Student & Teacher & Student & Teacher & Student & Teacher & Student & Teacher & Student\\
\hline
 1   &  76.0 $\pm$ 3.4 & 75.0 $\pm$ 4.0 & 76.6 $\pm$ 2.9 & 77.7 $\pm$ 2.0 & \textbf{78.5 $\pm$ 6.4} & 77.4 $\pm$ 5.1 & 42.2 & 42.3 & 49.6 & 49.2 & 54.3 & \textbf{55.0}\\
 5   &  74.6 $\pm$ 4.2 & 75.6 $\pm$ 4.2 & 76.8 $\pm$ 3.0 & 76.3 $\pm$ 3.7 & \textbf{81.7 $\pm$ 3.2} & 81.0 $\pm$ 4.0 & 45.3 & 45.7 & 53.3 & 53.2 & \textbf{56.1} & 55.8\\
 10  &  76.3 $\pm$ 4.1 & 75.6 $\pm$ 4.6 & 76.3 $\pm$ 5.0 & 76.0 $\pm$ 5.2 & 77.8 $\pm$ 2.8 & \textbf{77.9 $\pm$ 3.4} & 47.2 & 46.3 & 54.3 & 54.5 & 56.4 & \textbf{56.7}\\
 20  &  73.9 $\pm$ 3.5 & 73.9 $\pm$ 3.5 & 75.7 $\pm$ 5.3 & 72.9 $\pm$ 5.8 & 76.3 $\pm$ 3.0 & \textbf{76.5 $\pm$ 4.0} & 48.2 & 47.6 & 53.3 & 51.5 & \textbf{58.0} & 57.6\\
 50  &  73.5 $\pm$ 4.3 & 73.0 $\pm$ 4.1 & 74.2 $\pm$ 5.1 & 73.4 $\pm$ 6.5 & \textbf{74.7 $\pm$ 4.7} & 74.3 $\pm$ 5.7 & 47.0 & 47.3 & 50.8 & 49.7 & \textbf{57.5} & 56.9\\
 100 &  72.8 $\pm$ 3.7 & 73.0 $\pm$ 3.1 & 73.6 $\pm$ 5.8 & 71.4 $\pm$ 7.4 & \textbf{74.0 $\pm$ 5.5} & 73.6 $\pm$ 5.9 & 45.5 & 45.0 & 48.4 & 48.1 & \textbf{54.0} & 53.7\\
\hline
\end{tabular}}
\vspace{0.2cm}
\label{table:pe_knn_bracs_bach_s8}
\end{table*}

\begin{table}[ht]
\centering
\caption{\textbf{A weighted $k$ Nearest Neighbors classifier assesses the discriminability (weighted F1-score) of the learned features} on the tile CAMELYON16 dataset (see Sec.~\ref{subsec:dataset}). A fixed and absolute PE (2D sine-cos)'s performances are compared to a learnable and conditional PE (CPVT and CPVT-GAP). A 4-fold approach with $75\%/25\%$ train/test splits is used.}
\resizebox{\columnwidth}{!}{
\begin{tabular}{l c c c c c c}
\hline
       & \multicolumn{2}{c}{2D sine-cos} & \multicolumn{2}{c}{CPVT} & \multicolumn{2}{c}{CPVT-GAP} \\
\cmidrule(lr){2-3} \cmidrule(lr){4-5} \cmidrule(lr){6-7} 
$k$  & Teacher & Student & Teacher & Student & Teacher & Student \\
\hline
 1   & 88.2 $\pm$ 0.9 & 88.4 $\pm$ 0.6 & 88.8 $\pm$ 0.4 & 88.8 $\pm$ 0.4 & \textbf{89.1 $\pm$ 0.4} & 89.1 $\pm$ 0.4\\
 5   & 91.1 $\pm$ 0.9 & 91.0 $\pm$ 1.0 & 90.9 $\pm$ 0.5 & 90.9 $\pm$ 0.6 & 91.1 $\pm$ 0.5 & \textbf{91.2 $\pm$ 0.6}\\
 10  & 91.2 $\pm$ 0.7 & 91.2 $\pm$ 0.5 & 91.1 $\pm$ 0.4 & 91.1 $\pm$ 0.3 & 91.4 $\pm$ 0.5 & \textbf{91.5 $\pm$ 0.4}\\
 20  & 91.1 $\pm$ 0.7 & 91.1 $\pm$ 0.6 & 91.3 $\pm$ 0.5 & \textbf{91.4 $\pm$ 0.5} & 91.2 $\pm$ 0.3 & 91.2 $\pm$ 0.4\\
 50  & 90.5 $\pm$ 0.7 & 90.6 $\pm$ 0.7 & \textbf{90.8 $\pm$ 0.6} & \textbf{90.8 $\pm$} 0.5 & 90.7 $\pm$ 0.6 & 90.7 $\pm$ 0.5\\
 100 & \textbf{90.2 $\pm$ 1.8} & \textbf{90.2 $\pm$ 0.8} & \textbf{90.2 $\pm$ 0.6} & \textbf{90.2 $\pm$ 0.6} & 90.1 $\pm$ 0.6 & 90.0 $\pm$ 0.5\\
\hline
\end{tabular}
}
\vspace{0.2cm}
\label{table:pe_knn_camelyon_128}
\end{table}


\noindent\textbf{Selecting the Number of Finely Attended Regions.}
\label{par:selecting_k}The effect of the number of selected regions is depicted in Table ~\ref{table:selecting_k}. One can observe that it does not appear as the most determining factor, particularly that the results are not monotonically increasing, which is unexpected. There are two potential explanations for this behavior. The first is due to the heterogeneity of the BRACS dataset. More precisely, it encompasses images containing less than 50 patches, which implies that the image must first be resized, potentially harming the predictions. The second explanation is that the model used for this ablation is a $\mathrm{ScoreNet/4/1}$ variant, which by design relies less on the high-resolution images than its $\mathrm{ScoreNet/4/3}$ counterpart. The respective properties of these two variants are subject to Sec.~\ref{subsec:scorenet_properties}.
\begin{table*}[ht]
\centering
\caption{\textbf{The number of finely attended regions is selected} by independently training our pipeline $5$ times on $10\%$ of the BRACS dataset with a varying number of proposal regions. The number of training epochs is fixed and is the same for all experiments. The models are trained with standard data augmentation methods, i.e., none of ScoreMix, SaliencyMix, or CutMix.}
\resizebox{1\textwidth}{!}{
\begin{tabular}{c c c c c c c c | c}
\hline
\# Regions & Normal & Benign & UDH & ADH & FEA & DCIS & Invasive & Weighted F1  \\
\hline
$k=5$  & \textbf{53.7 $\pm$ 5.2} & \textbf{44.0 $\pm$ 5.1} & 29.7 $\pm$ 5.3 & 28.8 $\pm$ 6.8 & 69.3 $\pm$ 4.2 & 56.9 $\pm$ 6.5 & \textbf{86.9 $\pm$} 3.2 & 54.2 $\pm$ 1.8 \\
$k=10$ & 52.1 $\pm$ 6.2 & \textbf{44.0 $\pm$ 3.9} & \textbf{31.0 $\pm$ 5.3} & 28.6 $\pm$ 4.3 & 69.8 $\pm$ 3.6 & 56.4 $\pm$ 3.9 & 85.9 $\pm$ 1.4 & 54.0 $\pm$ 0.8 \\
$k=20$ & 52.2 $\pm$ 3.4 & 42.2 $\pm$ 5.6 & 29.6 $\pm$ 7.5 & \textbf{31.9 $\pm$ 5.3} & \textbf{71.9 $\pm$ 2.3} & \textbf{57.5 $\pm$ 3.6} & \textbf{86.9 $\pm$ 2.5} & \textbf{54.7 $\pm$ 0.8} \\
$k=50$ & 51.5 $\pm$ 5.4 & 42.8 $\pm$ 4.7 & 30.0 $\pm$ 6.8 & 25.9 $\pm$ 7.1 & 70.5 $\pm$ 4.0 & 55.8 $\pm$ 5.2 & 85.7 $\pm$ 0.9 & 53.3 $\pm$ 2.5 \\
\hline
\end{tabular}
}
\label{table:selecting_k}
\end{table*}


\subsection{ScoreNet Under the Magnifying Glass}
\label{subsec:scorenet_properties}

 \noindent\textbf{Just a Glorified Low-resolution ViT?}
\label{par:glorified_vit}
 We explore the usage of high-resolution images for predictions. For that purpose, at test time, we mask $75\%$ of the selected high-resolution regions and report the obtained results in Table~\ref{table:masking_ablation}. As expected, we observe that the $\mathrm{ScoreNet/4/3}$ variant uses the high-resolution content more. Furthermore, these results shed light on how the high-resolution information is not equally relevant for each class. An interesting observation is that for each variant of $\scorenet$, the higher the performance of a given model is, the more it is affected by the removal of the high-resolution information (see Table~\ref{table:masking_ablation2}). 
\begin{table*}[ht]
\centering
\caption{\textbf{At test time, $75\%$ of the selected high-resolution regions are randomly masked.} $\mathrm{ScoreNet/4/1}$ and $\mathrm{ScoreNet/4/3}$ do not equally rely on the high-resolution content.
}
\resizebox{1\textwidth}{!}{
\begin{tabular}{l c c c c c c c | c}
\hline
Masking & Normal & Benign & UDH & ADH & FEA & DCIS & Invasive & Weighted F1  \\
\hline
$\mathrm{ScoreNet/4/1}$ & 64.6 $\pm$ 2.2 & 52.6 $\pm$ 2.8 & 48.4 $\pm$ 2.2 & 47.4 $\pm$ 2.4 & 77.9 $\pm$ 0.7 & 59.3 $\pm$ 1.1 & 90.6 $\pm$ 1.5 & 64.1 $\pm$ 0.7 \\
Masked $\mathrm{ScoreNet/4/1}$ & 61.1 $\pm$ 2.7 & 50.8 $\pm$ 1.4 & 45.9 $\pm$ 2.2 & 41.0 $\pm$ 3.5 & 78.8 $\pm$ 0.5 & 59.9 $\pm$ 3.3 & 90.6 $\pm$ 1.1 & 62.4 $\pm$ 0.6 \\
\hdashline
$\mathrm{ScoreNet/4/3}$ & 64.3 $\pm$ 1.5 & 54.0 $\pm$ 2.2 & 45.3 $\pm$ 3.4 & 46.7 $\pm$ 1.0 & 78.1 $\pm$ 2.8 & 62.9 $\pm$ 2.0 & 91.0 $\pm$ 1.4 & 64.4 $\pm$ 0.9 \\
Masked $\mathrm{ScoreNet/4/3}$ & 64.9 $\pm$ 2.4 & 51.7 $\pm$ 0.5 & 44.4 $\pm$ 4.0 & 22.0 $\pm$ 6.2 & 77.6 $\pm$ 1.0 & 60.8 $\pm$ 1.6 & 87.2 $\pm$ 1.3 & 59.6 $\pm$ 0.7 \\
\hline
\end{tabular}}
\label{table:masking_ablation}
\end{table*}

\begin{table}[H]
\centering
\caption{\textbf{The performance drop incurred by the high-resolution masking operation of individual models is monitored}. The models that rely the most on the high-resolution content are the ones that perform the best.
}
\begin{tabular}{c c c | c c c}
\hline
       \multicolumn{3}{c}{$\mathrm{ScoreNet/4/1}$} & \multicolumn{3}{c}{$\mathrm{ScoreNet/4/3}$} \\
\cmidrule(lr){1-3} \cmidrule(lr){4-6} 
 63.3 & $\xrightarrow[]{-0.6}$ & 62.7 $\pm$ 0.2 & 63.3 & $\xrightarrow[]{-2.8}$ & 60.5 $\pm$ 0.1\\
 63.8 & $\xrightarrow[]{-2.2}$ & 61.6 $\pm$ 0.1 & 64.8 & $\xrightarrow[]{-5.2}$ & 59.6 $\pm$ 0.3\\
 64.9 & $\xrightarrow[]{-2.2}$ & 62.7 $\pm$ 0.3 & 65.0 & $\xrightarrow[]{-6.4}$ & 58.6 $\pm$ 0.3\\
\hline
\end{tabular}
\label{table:masking_ablation2}
\end{table}


		

 \begin{table*}[h]
\centering
\caption{\textbf{
The ViT network of recommendation stage is trained without receiving any feedback from the high-resolution-based predictions.} Its features discriminability is significantly worth than that of the same model but trained jointly with the high-resolution stage.
}
\resizebox{1\textwidth}{!}{
\begin{tabular}{l c c c c c c c | c}
\hline
Model & Normal & Benign & UDH & ADH & FEA & DCIS & Invasive & Weighted F1  \\
\hline
  $\vit$ & 53.3 $\pm$ 2.8 & 42.8 $\pm$ 1.9 & 37.1 $\pm$ 2.9 & 32.4 $\pm$ 2.4 & 77.3 $\pm$ 0.2 & 51.2 $\pm$ 1.3 & 85.0 $\pm$ 1.8 & 55.5 $\pm$ 0.1 \\
  Lin. scorer's [CLS] & 57.5 $\pm$ 4.2 & 48.8 $\pm$ 5.5 & 42.7 $\pm$ 3.5 & 42.7 $\pm$ 7.4 & 74.3 $\pm$ 5.2 & 60.5 $\pm$ 2.4 & 90.6 $\pm$ 0.2 & 60.9 $\pm$ 3.1 \\
  \hdashline
  $\mathrm{ScoreNet/4/1}$ & \textbf{64.6 $\pm$ 2.2} & 52.6 $\pm$ 2.8 & \textbf{48.4 $\pm$ 2.2} & \textbf{47.4 $\pm$ 2.4} & 77.9 $\pm$ 0.7 & 59.3 $\pm$ 1.1 & 90.6 $\pm$ 1.5 & 64.1 $\pm$ 0.7 \\
  $\mathrm{ScoreNet/4/3}$ & 64.3 $\pm$ 1.5 & \textbf{54.0 $\pm$ 2.2} & 45.3 $\pm$ 3.4 & 46.7 $\pm$ 1.0 & \textbf{78.1 $\pm$ 2.8} & \textbf{62.9 $\pm$ 2.0} & \textbf{91.0 $\pm$ 1.4} & \textbf{64.4 $\pm$ 0.9} \\
\hline
\end{tabular}}
\vspace{0.2cm}
\label{table:glorified_vit}
\end{table*}

 Despite that, we expected a more considerable drop in performance from this masking operation, which raises the question; \textit{is $\scorenet$ nothing but a glorified low-resolution $\vit$?} To answer that question, we train the same $\vit$ as the one used in the recommendation stage and the same setting, but basing the predictions on the scorer's [CLS] tokens and hence without the feedback from the high-resolution stage. Table~\ref{table:glorified_vit} clearly shows a gap of almost $10\%$ compared to $\scorenet$'s results and, more interestingly, a gap of more than $5\%$ when compared to the same $\vit$, but trained with the high-resolution feedback. The above results indicate that \textbf{ high-resolution information distillation occurs during the training of $\snet$}.

\section{Computational Cost}
\label{sec:comp_cost}
Vision transformers heavily rely on the attention mechanism to learn a high-level representation from low-level regions. The underlying assumption is that the different sub-regions of the image are not equally important for the overall representation. Despite this key observation, the computation cost dedicated to a sub-region is independent of its contribution to the high-level representation, which is inefficient and undesirable. Our $\scorenet$ attention mechanism overcomes this drawback by learning to attribute more computational resources to regions of high interest. Let us consider a high-resolution input image $x_{h} \in \mathbb{R}^{C \times H \times W}$, a low-resolution version of the image $x_{l} \in \mathbb{R}^{C \times h \times w}$ is obtained by applying a down-scaling factor $s$, as $h = H/s$ and $w = W/s$. The low-resolution image is fed to a scorer model (recommendation stage), which recommends the regions where to apply fine-grained attention. If this operation is implemented by a $\vit$, its computational cost is $\mathcal{O} \left(\left( \frac{h}{P_{l}} \cdot \frac{w}{P_{l}}\right)^2\right)$ with $P_{l}$ is the dimension of the patches in low-resolution. Using a $\vit$ as the scorer model, there is a one-to-one mapping between the low-resolution patches and the regions the model can process with fine-grained attention; as a consequence, the dimension of the regions is $P_{h} = s \cdot P_{l}$. Attending to such regions with a patch size, $P_{a}$, has a computational cost of $\mathcal{O} \left(\left( \frac{P_{h}}{P_{a}} \cdot \frac{P_{h}}{P_{a}} \right)^2\right)$ and the model processes $k$ of them, hence $\mathcal{O} \left( k \cdot \left(\frac{P_{h}}{P_{a}} \cdot \frac{P_{h}}{P_{a}} \right)^2\right)$. Finally, a coarse attention mechanism is applied to endow the locally attended regions with contextual information. This final step costs $\mathcal{O} \left( k^2 \right)$. On the other hand, a vanilla ViT would attend uniformly across the whole image with a cost of $\mathcal{O} \left(\left( \frac{H}{P_{a}} \cdot \frac{W}{P_{a}}\right)^2\right)$. Importantly, we observe that only the recommendation stage's cost depends on the input size; consequently, if this step is implemented as a $\vit$ and with a down-scaling ratio $s \in [8, 14]$, the asymptotic cost is reduced by approximately two orders of magnitude, as we typically used $P_{a} = P_{l}$ in practice. At last, one can observe that the asymptotic cost can be made linear w.r.t. the input dimension by adopting a convolution-based architecture for the recommendation stage.

\section{ScoreMix Investigation \& Examples}
\label{sec:scoremix_investigation}
\begin{figure*}[ht]
    \centering
    \includegraphics[width=0.85\textwidth]{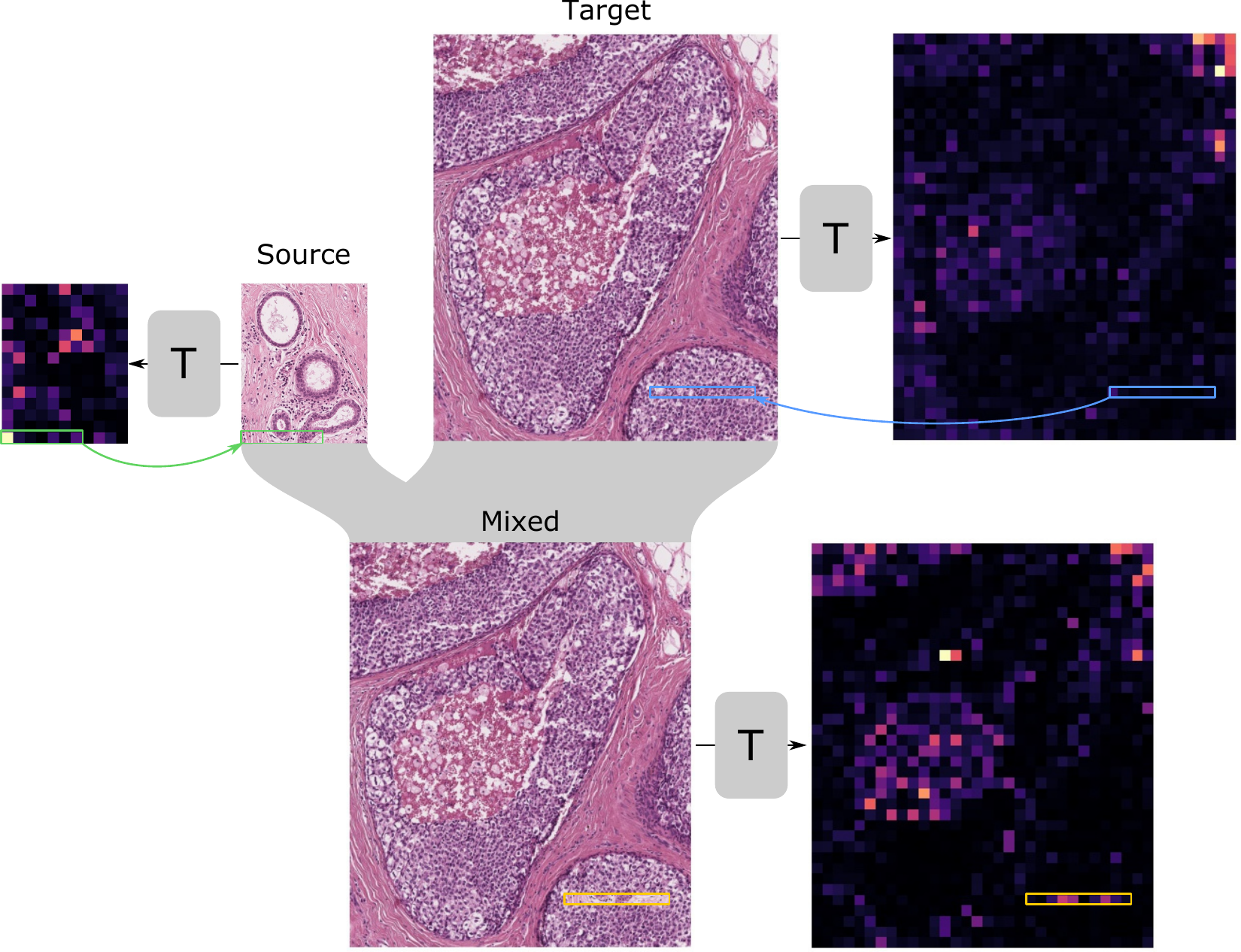}
    \caption{\textbf{The learned semantic distribution can detect and localize the newly pasted content.} The green box highlights the region pasted from the source to the target image. The blue box represents the region where the new content is pasted. The yellow box highlights the modified region in the mixed image. \textit{T} represents the scorer network of $\scorenet$.}
    \label{fig:scoremix_example}
\end{figure*}

The underlying assumption of the ``cut-and-paste''-based augmentation methods is that the trained model can assimilate the pasted region to the representation of the target image. In the case of $\scorenet$, it translates to attending to the pasted area in a low or high-resolution image. Fig.~\ref{fig:scoremix_example} depicts an example of $\scorenet$ being able to detect and localize the pasted regions even when the pasted region is small and hard to distinguish. We further observe that a local change in the image directly affects the global representations as the representation of each token is adapted to accommodate the local change in information. This behavior would typically not be observed in a CNN-based architecture until the very last layers. Fig.~\ref{fig:scoremix_example} further highlights the ability of $\scoremix$ to treat images of different dimensions and aspect ratios.

\section{Learning From Uncurated Data.} 
\label{sec:uncurated_learning}
We gauge the ability of $\scorenet$ to learn from unlabeled data on the BACH dataset \cite{ARESTA2019122}, which encompasses both a small set of $400$ annotated $\trois$ images, and the $\wsis$ containing the aforementioned $\trois$. Our model is first pre-trained using DINO's self-supervised learning scheme~\cite{caron2021emerging} on an unlabeled set of $\approx11$k images extracted from $\wsis$ and then is evaluated on the labeled image set using standard protocols, namely linear probing and $k$-NN (see Table~\ref{table:bach_knn}). We also report the non-empty cluster's purity for the clusters learned by DINO. This metric indicates the quality of a cluster containing samples from a single class. Learning from large uncurated images is particularly challenging, as the increased receptive field allows for the representation of more complex tissue interactions. This further deviates from the discriminative pretext task's assumption that the images represent a single centered object. Since the DINO method enforces a local-to-global correspondence between large and smaller image crops, it may enforce similarity between different tissue types. For that purpose, we modify DINO's pretext task so that the student network only processes the highly discriminative patches to match the teacher's representation, allowing the processing of all the high-resolution patches. To ensure that the pretext task does not encourage contextual biases \cite{singh2020don}, we only employ augmentations that change the image pixels' values, but not their locations, such that the semantic content of the two augmented views is identical. As can be observed in Table~\ref{table:bach_knn}, this proposed strategy yields significant improvements compared to other baselines. 
\begin{table}[H]
\centering
\caption{\textbf{Comparison with the prior art for learning capabilities from uncurated data} on the BACH dataset using DINO's pre-training. A comparison results between the effectiveness of DINO's standard pretext task ($\scorenet$) and the proposed unbiased pretext task ($\scorenet^{\dagger}$) are also reported.}
\resizebox{\columnwidth}{!}{
\begin{tabular}{ l c c c c}
\hline
       & \cellcolor{violet!20} ScoreNet$^{\dagger}$ & \cellcolor{violet!20} ScoreNet & TransPath \cite{wang2021transpath} & SwinTransformer \cite{liu2021swin} \\
\hline
$k$-NN       & \cellcolor{violet!20} \textbf{73.7 $\pm$ 1.7} & \cellcolor{violet!20} 65.0 $\pm$ 3.7 & 65.2 $\pm$ 1.4 & 63.7 $\pm$ 4.1 \\
Lin. eval    & \cellcolor{violet!20} \textbf{73.0 $\pm$ 2.9} & \cellcolor{violet!20} 66.0 $\pm$ 2.6 & 64.2 $\pm$ 4.0 & 62.5 $\pm$	1.7  \\
Purity       & \cellcolor{violet!20} \textbf{78.3 $\pm$ 23.9} & \cellcolor{violet!20} 76.4 $\pm$ 24.9 & 74.0 $\pm$ 23.3 & 71.8 $\pm$ 23.9 \\
\hline
\end{tabular}
}
\vspace{0.2cm}
\label{table:bach_knn}
\end{table}

\end{document}